\pdfoutput=1

\documentclass[11pt]{article}

\usepackage[preprint]{acl}

\usepackage{times}
\usepackage{latexsym}

\usepackage[T1]{fontenc}

\usepackage[utf8]{inputenc}

\usepackage{microtype}

\usepackage{inconsolata}

\usepackage{graphicx}

\usepackage{amsmath}
\usepackage{cleveref}
\definecolor{darkgreen}{rgb}{0.0, 0.5, 0.0}

\newcommand{\bench}{\textsc{\textbf{F}\textbf{i}na\textbf{l}}\xspace}

\newcommand{\enumlabel}[1]{\textbf{\textcolor{teal!70!black}{(#1)}}}

\usepackage{array}
\usepackage{tabularx}
\usepackage{booktabs}
\usepackage{pifont}
\usepackage{xcolor}
\usepackage{adjustbox}
\usepackage{multirow}
\usepackage{placeins}
\usepackage{soul}
\usepackage{enumitem}
\usepackage{float}
\restylefloat{table}
\usepackage{tikz}
\usepackage{ragged2e}
\usepackage[most]{tcolorbox}
\usepackage{pgfplots}
\usepackage{caption}
\usepgfplotslibrary{groupplots} 
\pgfplotsset{compat=1.18}
\usepgfplotslibrary{colorbrewer}
\usepackage{pgfplotstable} 
\usepackage[dvipsnames]{xcolor}
\usetikzlibrary{positioning, arrows.meta, shapes.misc}
\usetikzlibrary{tikzmark,decorations.pathreplacing,calc}
\usepackage{setspace}
\usepackage[normalem]{ulem}
\usepackage{makecell}
\usepackage{xspace}
\usepackage{microtype}

\title{Fine‑Grained Detection of Context‑Grounded Hallucinations Using LLMs}

\author{
  Yehonatan Peisakhovsky$^{\text{T*}}${\hspace{.1em}} \quad
  Zorik Gekhman$^{\text{T*}}${\hspace{.1em}} \quad
  \textbf{Yosi Mass}$^{\text{I}}${\hspace{.1em}}  \quad
  \textbf{Liat Ein-Dor}$^{\text{I}}${\hspace{.1em}}   \quad
  \textbf{Roi Reichart}$^{\text{T}}${\hspace{.1em}}  
  \vspace{1.1em}\\
  $^*$ Equal contribution; author order was chosen randomly.
  \vspace{0.3em}\\
  \textsuperscript{T}Technion - Israel Institute of Technology \quad \textsuperscript{I}IBM Research
  \vspace{0.3em}\\
  yonip1997@gmail.com \quad
  zorikgekhman@gmail.com \quad
  roiri@technion.ac.il
}

\begin{document}
\maketitle
\begin{abstract}

Context-grounded hallucinations are cases where model outputs contain information not verifiable against the source text.
We study the applicability of LLMs for \textit{localizing} such hallucinations, as a more practical alternative to existing complex evaluation pipelines.
In the absence of established benchmarks for \textit{meta-evaluation} of hallucinations \textit{localization}, we construct one tailored to LLMs, involving a challenging human annotation of over 1,000 examples.
We complement the benchmark with an LLM-based evaluation protocol, verifying its quality in a human evaluation.
Since existing \textit{representations} of hallucinations limit the types of errors that can be expressed, we propose a new representation based on free-form textual descriptions, capturing the full range of possible errors.
We conduct a comprehensive study, evaluating four large-scale LLMs, which highlights the benchmark's difficulty, as the best model achieves an F1 score of only 0.67.
Through careful analysis, we offer insights into optimal prompting strategies for the task and identify the main factors that make it challenging for LLMs: (1) a tendency to incorrectly flag missing details as inconsistent, 
despite being instructed to check only facts in \textit{the output};
and (2) difficulty with outputs containing factually correct information absent from the source - and thus not verifiable - due to alignment with the model's parametric knowledge.

\end{abstract}

\newcommand{\myfont}{\fontsize{9pt}{12pt}\selectfont}
\definecolor{color1}{RGB}{200, 240, 200} 
\definecolor{color2}{RGB}{180, 210, 255} 

\newtcolorbox{whitebox}[1][]{%
  colback   = white,         
  colframe  =gray,    
  coltitle  = black,  
  fonttitle =  \small\bfseries,   
fontupper = \myfont, 
  title  = #1,                 
  enhanced,                       
  attach boxed title to top center = {yshift=-2mm},
  boxed title style = {
    colback = white,
    arc = 3pt,
    boxrule = 0.8pt,
       top=0pt,    
  bottom=0pt, 
  left=2pt,    
  right=2pt  
  },
  arc = 3pt,                
  boxrule = 0.8pt,              
  left = 3pt, right = 3pt,
  top  = 6pt, bottom = 0pt        
}

\newtcbox{\firstcolorbox}{on line,
  box align=base,
  fontupper = \myfont, 
  colback=color1,
  colframe=white,
  boxrule=0pt,
  arc=2pt, 
  left=0pt, right=0pt, top=0pt, bottom=0pt,
  boxsep=1pt,
  enhanced,
  nobeforeafter
}

\newtcbox{\secondcolorbox}{on line,
  box align=base,
  fontupper = \myfont, 
  colback=color2,
  colframe=white,
  boxrule=0pt,
  arc=2pt, 
  left=0pt, right=0pt, top=0pt, bottom=0pt,
  boxsep=1pt,
  enhanced,
  nobeforeafter
}

\begin{figure}[ht]
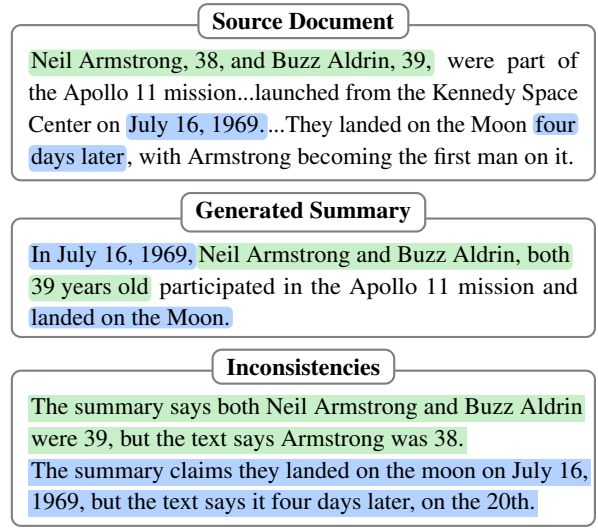

  \centering

  \begin{minipage}{\columnwidth}
\begin{whitebox}[Source Document]
\firstcolorbox{Neil Armstrong, 38, and Buzz Aldrin, 39,} were part of the Apollo 11 mission...launched from the Kennedy Space Center on \secondcolorbox{July 16, 1969.}...They landed on the Moon \secondcolorbox{four} \secondcolorbox{days later}, 
with Armstrong becoming the first man on it.
\end{whitebox}
  \end{minipage}
  
\vspace{3pt}

\begin{minipage}{\columnwidth}
  \begin{whitebox}[Generated Summary]
  \secondcolorbox{In July 16, 1969,}%
  \firstcolorbox{ Neil Armstrong and Buzz Aldrin, both}
  \firstcolorbox{39 years old} participated in the Apollo 11 mission and \secondcolorbox{landed on the Moon.}
  \end{whitebox}
\end{minipage}

\vspace{3pt}

  \begin{minipage}{\columnwidth}
    \begin{whitebox}[Inconsistencies]
{\setlength{\fboxsep}{1pt}\colorbox{color1}{The summary says both Neil Armstrong and Buzz Aldrin}}
{\setlength{\fboxsep}{1pt}\colorbox{color1}{were 39, but the text says Armstrong was 38.}}
{\setlength{\fboxsep}{1pt}\colorbox{color2}{The summary claims they landed on the moon on July 16,}}
{\setlength{\fboxsep}{1pt}\colorbox{color2}{1969, but the text says it four days later, on the 20th.}}

    \end{whitebox}
  \end{minipage}
  \vspace{-2pt}
  \caption{
  Since factual inconsistencies can be complex and hard to represent, we propose a representation based on free-form textual descriptions in natural language.}
  \label{fig:source-summary}
  \vspace{-5pt}
\end{figure}

\section{Introduction}
The ability to generate responses conditioned explicitly on a given input is critical for many downstream tasks, including summarization \citep{pmlr-v119-zhang20ae}, open book question-answering \citep{nakano2021webgpt} and retrieval-augmented generation \citep{OriginalRAG}.
In such context-grounded setups, a response is considered \textit{factually consistent} 
if any piece of information it contains
is supported by the source
text \citep{bohnet2022attributed}.

Early work on factual consistency \textit{evaluation} used a \textit{binary} setup, classifying the entire output as consistent or not \citep{True,Summac, trueteacher}. Such binary classification not only overlooks severity by scoring outputs with many errors the same as those with only one, but also fails to \textit{localize} hallucinations (pinpointing which parts of the output are not supported): a critical capability for analyzing model failure patterns and enabling targeted corrections. This motivated a shift to \mbox{\textit{fine-grained}} evaluation, analyzing smaller output units such as entities \citep{cao2022}, spans \citep{Colm_fine_grain_detection_and_edit}, question-answer (QA) pairs \citep{q2}, or atomic facts \citep{factscore}. 

While the shift to fine-grained evaluation marks an important step towards more reliable evaluation, existing methods still have fundamental limitations.
First \enumlabel{i} the \textit{error-representation} in prior work using formats such as entities, spans, QA pairs, and atomic facts, constrain the error types that can be captured, making it infeasible to represent the full spectrum of factual errors (Table~\ref{tab:annotation_differences} and \S\ref{sec:representation}).
Second \enumlabel{ii} most methods output a continuous score, rather than identifying specific errors \citep{factscore}, which limits their usefulness for error localization.

The third limitation \enumlabel{iii} is that existing methods are often based on complex, multi-stage pipelines that can be difficult to train, maintain, and deploy effectively, limiting practical usability. A promising, simpler alternative is to leverage large language models (LLMs) for end-to-end localization. However, the extent to which LLMs can reliably localize factual inconsistencies remains unclear.\footnote{LLMs are used for intermediate steps in pipeline-based methods \citep{factscore}, but not for end-to-end evaluation.} 
Lastly \enumlabel{iv} the lack of established frameworks for reproducible automatic \textit{meta-evaluation} makes it challenging to compare different evaluation systems. 
This limitation is even more pronounced when using LLMs, given their rapid development and wide variety.

We argue that existing evaluation pipelines should be replaced by LLMs, increasing practical usability \enumlabel{iii}. To support automatic evaluation \enumlabel{iv}, we curate the \bench benchmark for evaluating LLMs on the task of \textbf{F}actual \textbf{I}nconsistencies \textbf{L}ocalization.
It is constructed from partially annotated examples in the \mbox{DeFacto} dataset \citep{DeFacto}, with the curation involving a challenging human annotation task of over 1,000 examples.

To capture the full range of errors \enumlabel{i}, we represent inconsistencies as free-form natural language \textit{descriptions}, which provide maximal flexibility and align naturally with the strengths of LLMs.
This choice also helps to localize errors \enumlabel{ii}, as the LLM generates a list of interpretable descriptions rather than a single continuous score.
To enable the evaluation of such lists of errors, we design an LLM-based evaluation protocol that complements our benchmark, and whose judgment quality is validated through human evaluation.

We show that our benchmark is challenging even for strong LLMs by evaluating four large-capacity models using various prompting strategies: Llama-3-405B, GPT-4o, Gemini-Pro, and Claude-Sonnet, with the best-performing model achieving an F1 score of 0.67. 
Our analysis shows that (1) reasoning aids in addressing this task
and (2) 
two-step approaches, where the model first classifies the summary as consistent or inconsistent and then identifies individual errors, tend to perform worse than end-to-end localization due to a conservative behavior in the binary step, leading to low recall.

Lastly, we conduct a comprehensive error analysis 
grouping localization failures into categories with interpretable meanings, 
which allows us to identify two key weaknesses: (1) LLMs tend to incorrectly treat missing information in the output as inconsistent, even when explicitly instructed to examine only facts present in the output, and (2) LLMs struggle when the output has correct information not present in the source text: although such information cannot be verified by the source, its alignment with the model's parametric knowledge causes the model to misclassify it as consistent.
To summarize, our contributions are as follows:

\vspace{-5pt}
\begin{itemize}
    \item We create a benchmark for the meta-evaluation of LLMs on the task of fine-grained factual consistency evaluation.\footnote{\href{https://github.com/yonip97/The_final_benchmark}{github.com/yonip97/The\_final\_benchmark}} 
    \vspace{-5pt}

    \item We propose a new paradigm for error representation based on free-form textual description, which allows to represent \textit{any} possible error.
    \vspace{-5pt}

    \item We design an LLM-based evaluation protocol for error localization and validate its quality in a human evaluation. 
    \vspace{-5pt}

    \item We conduct a comprehensive evaluation of four large-scale LLMs on the task, exploring different prompting strategies and comparing the end-to-end and two-step paradigms.
    \vspace{-5pt}
    \item We conduct a comprehensive error analysis, group failures into meaningful categories, and uncover key weaknesses of LLMs on the task.
    \vspace{-5pt}

\end{itemize}
\vspace{-5pt}
\vspace{-10pt}
\section{Representing Factual Inconsistencies via Descriptions in Natural Language}
\label{sec:representation}

\newcommand{\highlightword}[3]{
  \tikz[baseline=(X.base)]{
    \node[inner sep=0pt, outer sep=0pt, anchor=base west] (X) at (0,0) {\strut#1};
    \definecolor{tempHL}{RGB}{#3}
    \fill[tempHL] 
      (X.north west) 
      rectangle 
      ([yshift=-#2]X.north east);
    \node[inner sep=0pt, outer sep=0pt, anchor=base west] at (0,0) {\strut#1};
  }%
}
\newcommand{\cmark}{\textcolor{green}{\ding{51}}} 
\newcommand{\xmark}{\textcolor{red}{\ding{55}}}
\begin{table*}[t]
    \small 
    \resizebox{\textwidth}{!}{%
    \begin{tabular}{p{2.75cm} p{16cm}} 
        \toprule
        \textbf{Text} & 
         After surpassing 250,000 units sold on Amazon, Philips attracted the attention of the authorities, prompting an investigation... After investigating for a month, the police concluded that Philips could continue operating without restrictions.
        \\
        \midrule
        \textbf{Summary} & Following a record number of sales, Amazon was investigated for a month and ordered by the police to cease operations.\\
        \midrule
        \textbf{Entities} &Following a record number of sales,{\setlength{\fboxsep}{1pt}\colorbox[RGB]{182, 215, 168}{Amazon}} was investigated for a month and ordered by the police to cease operations. \\
        \midrule
        \textbf{Spans} &Following a
\highlightword{record}{2ex}{191,228,255}%
\highlightword{ number}{1.2ex}{191,228,255}%
\highlightword{ of}{0.6ex}{191,228,255}%
\highlightword{ sales}{0.4ex}{191,228,255}, 
{\setlength{\fboxsep}{1pt}\colorbox[RGB]{182, 215, 168}{Amazon}} was investigated for a month and {\setlength{\fboxsep}{1pt}\colorbox[RGB]{255, 229, 153}{ordered \textcolor{red}{by the police} to cease operations.}}
\\
        \midrule
 \textbf{Atomic Facts} & 
\begin{tabular}{@{}p{7.9cm}@{} @{}p{7.9cm}@{}}
    \textbullet\ Amazon made a record number of sales.\xmark
    & \textcolor{white}{\textbullet} \\
    
    \textbullet\ Amazon was investigated.\xmark
    &\tikzmark{target1} The police\\
    
    \textbullet\ The investigation lasted a month.\cmark
    & \tikzmark{target2} Amazon\\
    
    \textbullet\ The police ordered Amazon to cease operations.\xmark\tikzmark{source1}
    &\tikzmark{target3} Ordered to cease operations \\

\end{tabular}
\tikz[remember picture, overlay]{
  \draw[->, thick] ($(pic cs:source1)+(0.2em,0)$) -- ($(pic cs:target1)+(-0.2em,0.1)$);
  \draw[->, thick] ($(pic cs:source1)+(0.2em,0)$) -- ($(pic cs:target2)+(-0.2em,0.1)$);
  \draw[->, thick] ($(pic cs:source1)+(0.2em,0)$) -- ($(pic cs:target3)+(-0.2em,0.1)$);
}
\\
       \midrule
 \textbf{QA Pairs} & 
\begin{tabular}{@{}p{12cm}@{} @{}p{4cm}@{}}
   \textbullet\ What did {\setlength{\fboxsep}{1pt}\colorbox[RGB]{182, 215, 168}{Amazon}} accomplish? 
      {\setlength{\fboxsep}{1pt}\colorbox[RGB]{191, 228, 255}{A record number of sales}} \xmark
    & \textcolor{white}{\textbullet} \\
    
    \textbullet\ Who was investigated? 
     {\setlength{\fboxsep}{1pt}\colorbox[RGB]{182, 215, 168}{Amazon}} \xmark
    &\tikzmark{target4} Following \\
    
    \textbullet\ What was the police's action after the investigation? 
    {\setlength{\fboxsep}{1pt}{\colorbox[RGB]{255, 229, 153}{Order to cease operations}} }\xmark
    &\tikzmark{target5} Amazon \\
    
    \textbullet\ When was someone investigated? Following Amazon’s record number of sales \xmark \tikzmark{source2}
    &\tikzmark{target6} Record number \\
    
    \textbullet\ How long did the investigation take? A month \cmark
    &\tikzmark{target7} Sales  \\

    \textbullet\ Who investigated something? The police \cmark
    & \textcolor{white}{\textbullet} 
\end{tabular}
\tikz[remember picture, overlay]{
  \draw[->, thick] ($(pic cs:source2)+(0.2em,0)$) -- ($(pic cs:target4)+(-0.2em,0.1)$);
   \draw[->, thick] ($(pic cs:source2)+(0.2em,0)$) -- ($(pic cs:target5)+(-0.2em,0.1)$);
   \draw[->, thick] ($(pic cs:source2)+(0.2em,0)$) -- ($(pic cs:target6)+(-0.2em,0.1)$);
   \draw[->, thick] ($(pic cs:source2)+(0.2em,0)$) -- ($(pic cs:target7)+(-0.2em,0.1)$);
}
\vspace{-0.5em}
\\
        \midrule
     \textbf{Descriptions (Ours)} & 
    \parbox{15cm}{\begin{itemize}[leftmargin=*,noitemsep,topsep=0pt,parsep=0pt,partopsep=0pt]
    \item The summary calls the sales "a record," but the text says "surpassing 250,000 units" without mentioning a record.
    \item  The summary refers to Amazon, but the text says it was Philips being investigated.
    \item  The summary says the company was ordered to cease operations, but the text says it was authorized to continue.
    \end{itemize}} 
        \\
        \bottomrule
    \end{tabular}
    
    \caption{Examples of different strategies for annotating factual errors. 
    The summary has three errors: (1) the number of sales is not said to be a record, (2) the company is Philips, not Amazon, and (3) the company was not ordered to cease operations but was allowed to continue. In both Atomic Facts and QA pairs, all facts are explicitly listed, and the consistent and inconsistent ones are marked with \cmark and \xmark, respectively.
    }
    \label{tab:annotation_differences}
    }
    \vspace{-5pt}
\end{table*}

In this section we propose a new representation of errors.\footnote{For brevity, we use \textit{error} throughout the paper to refer specifically to factual inconsistencies.}
We first discuss the limitations of the representations from previous work (see \S\ref{sec:related}) using Table~\ref{tab:annotation_differences} as a running example.

\textit{Entities} limit coverage as they cannot represent errors in verbs, adjectives, general nouns, or more nuanced errors. For example, in Table~\ref{tab:annotation_differences} the summary claims a \textit{``record number of sales''}, while there is no evidence that a new record was set. Such error cannot be captured by highlighting an entity. 

\textit{Spans} can be subjective \citep{Colm_fine_grain_detection_and_edit}, as there are often many possible ways to annotate an error. For instance, in Table~\ref{tab:annotation_differences} we could highlight \textit{``record number of sales''}, \textit{``record number''} or \textit{``record''}. In addition, some errors do not correspond to a contiguous text sequence. E.g., in the final highlighted error the summary says \textit{``ordered by the police to cease operations''} while the text states that \textit{``Philips could continue operating without restrictions''}. Highlighting this span could be interpreted to mean that the order was not issued by the police.
Those issues introduce evaluation challenges, making it difficult to compare predicted inconsistencies against ground-truth annotations. Consequently, previous work resorted to simplified settings: sentence-level evaluation, which lacks granularity \cite{Colm_fine_grain_detection_and_edit}, and character-level span overlap, which is sensitive to minor shifts in span boundaries \cite{ragtruth}.

\textit{Atomic facts} and \textit{QA pairs} can be vague. For example, in Table \ref{tab:annotation_differences} the inconsistent label for the atomic fact \textit{``The police ordered Amazon to cease operations''}, does not indicate whether the error lies in the company name, the authority issuing the order, or the action itself. Another example is the QA pair \textit{''When was something investigated? Following Amazon's record number of sales.''}, which include several facts: the company (Amazon), the timing (after the record sales), and the outcome itself (record number of sales), making it unclear which part is inconsistent.
Moreover, generating questions from factually inconsistent summaries can cause those same factual errors to propagate into the questions \citet{qgqashortcomings}.

Lastly, we aim to create a benchmark for LLMs, and existing representations can be less natural for LLMs that output text in natural language.
We propose to use \textit{descriptions}: free-form explanations in natural language describing the nature of the error. While descriptions address the limited expressivity of existing representations, they introduce an evaluation challenge 
in comparing the model-generated descriptions to gold references. To address this, in \S\ref{Experimental setup Evaluation} we design an LLM-based evaluation protocol and verify its quality through human evaluation. 

\definecolor{customblue}{RGB}{0, 0, 255}  
\definecolor{customred}{RGB}{255, 0, 0}
\definecolor{customgreen}{RGB}{0, 160, 0}
\definecolor{customyellow}{RGB}{204, 172, 0}
\newcolumntype{L}[1]{>{\raggedright\arraybackslash}p{#1}}
\newcolumntype{S}[1]{>{\small\raggedright\arraybackslash}p{#1}}
\begin{table*}
\resizebox{\textwidth}{!}{
\begin{tabular}{
  L{2.2cm}  
  S{5.5cm}    
  S{3.5cm}   
  S{4.8cm}   
  S{5.5cm}   
  S{5cm}  
}
\toprule
\textbf{\large Type} & \textbf{\large Text} & \textbf{\large Summary} & \textbf{\large DeFacto Explanation} & \textbf{\large Our Descriptions} & \textbf{\large Comment} \\
\cmidrule(lr){1-6}
Extraction 
&Robin Clark, 44, was shot in the leg in the car park at Shenfield station ... has since returned to his job at RP Martin in London... a man from Essex has been arrested ... &A 46-year-old man has been arrested in connection with the shooting of a security guard at a London Underground station. & \textcolor{customblue}{No mention of his age}, \textcolor{customgreen}{that the other man was a security guard}, and \textcolor{customyellow}{it was not located at London Underground but in a park near Shenfield.}
& \parbox[t]{\linewidth}{
\textcolor{customblue}{No mention of his age in the source text.}\\[0.3em]
\textcolor{customgreen}{No mention that the other man was a security guard in the source text}\\[0.3em]
\textcolor{customyellow}{It was not located at London Underground but in a park near Shenfield.}
} & The individual inconsistencies are apparent in the original explanation. Only need to separate them into self contained factual inconsistency descriptions. \\
\midrule
Decomposition 
&Platt, 19 and Thomson, 21, have both joined the National League outfit until the end of the season... Blackburn are currently 22nd in the second tier... &Barrow have signed Blackburn Rovers midfielders Ben Platt and Josh Thomson on loan. & \textcolor{customblue}{Their first names are not mentioned} and second \color{customgreen}{it is not mentioned who signed them.} &\parbox[t]{\linewidth}{
\textcolor{customblue}{The first name of Platt is not in the  text.}\\[0.3em]
\textcolor{customblue}{The first name of Thomson is not in the text.}\\[0.3em]
\textcolor{customgreen}{It is not mentioned who signed them in the source text.}
} & The original explanation merges two distinct inconsistencies (2 first names) into a single description.  \\
\midrule
Vague Explanation
& ...complaint was made that police did not fully investigated claims against the Sinn Féin president...he had "found no evidence to indicate that [police officers] thinking was influenced by who Mr Adams was"...&Police in Northern Ireland have been cleared of any wrongdoing over their handling of allegations against Gerry Adams. &  \textcolor{customblue}{The summary incorrectly adds info about Northern Ireland} and \textcolor{customgreen}{Mr. Adams' first name.}& \parbox[t]{\linewidth}{
\textcolor{customblue}{The summary states it was Police in Northern Ireland, while the source text does not mention any location.}\\[0.3em]
\textcolor{customgreen}{The summary incorrectly adds Mr. Adams' first name.}} & The original explanation claims there's an issue with the information about Northern Ireland, but the actual inconsistency is that this location is not mentioned at all in the text. \\
\cmidrule(lr){1-6}
Missing Explanation 
&The 46-year-old number one seed defeated his 26-year-old opponent 7-3  ..."Now, thanks to hard work, determination and Teesside steel, I am world champion." ... &England's Martin Durrant has won his first BDO world title with victory over Australia's Scott Noppert. &There is info in the summary not found in the source, e.g. \textcolor{customblue}{BDO title}, etc. &\parbox[t]{\linewidth}{
\textcolor{customblue}{The summary calls it a "BDO world title," but the source doesn't name the organization.}\\[0.3em]
\textcolor{customgreen}{Noppert nationality is not in the source text
}} & The original explanation is lacking. It does not cover the inconsistency in the nationality of Noppert.\\
\midrule
Irrelevant Information  &The 35-year-old victim was attacked outside Barclays Bank ... The men, aged 41 and 42, were arrested on suspicion of murder...&Two men have been arrested after a man was stabbed outside a bank.&\textcolor{customred}{{Clearly states two men were arrested }} and \textcolor{customblue}{ the victim was attacked, but not necessarily stabbed.} &\textcolor{customblue}{The summary claims the man was stabbed, but the text only states he was attacked} &The original explanation contains information on why the summary is correct, not why it is inconsistent.\\
\midrule
Wrong Annotation 
& ... 89 out of 157 school closures between the academic years 2006-07 and 2015-16 were in the nine predominantly-rural council areas...&Almost half of school closures in Wales over the past decade were in rural areas, it has been claimed. & \textcolor{customred}{The source text does not say that September 2006 happened 10 years ago} or \textcolor{customblue}{that less than half of the closures were in rural areas.} &  \textcolor{customblue}{The source text does not say that less than half of the closures were in rural areas,but more than half.} & The explanation wrongly flags “the past decade” as incorrect, though the text supports it, so it is not an inconsistency. \\
\cmidrule(lr){1-6}
Fabricated 
&The victim was threatened with a knife and punched during the attack at Exhibition Park in the early hours. Her attacker is described as... Northumbria Police has... &A 19-year-old woman has been raped in Newcastle city centre. & It makes up the entire summary& & Summary is almost entirely fabricated and unrelated to the text, making fine-grained annotation meaningless, so those samples were excluded.\\
\bottomrule
\end{tabular}
}
\captionsetup{
  skip=0pt,
  belowskip=0pt
}
\caption{
Examples of annotation operations that were applied to convert the explanations from the DeFacto dataset \citep{DeFacto} to our error descriptions (\S \ref{sec:representation}).
}
\label{tab:manual_proccessing_annotation}
\end{table*}

\section{The \bench Benchmark}
\label{Benchmark}
To create the \bench benchmark for evaluating LLMs' performance on \textbf{F}actual \textbf{I}nconsistencies \textbf{L}ocalization, we need to (1) obtain a collection of source texts paired with corresponding outputs and (2) annotate them with factual errors as described in \S\ref{sec:representation}.
We chose to build on the \mbox{DeFacto} dataset \citep{DeFacto}, since it contains document-summary pairs that have already been partially annotated.
The goal of DeFacto was to explore how human feedback can help revise summaries, so it is annotated with \textit{explanations} for why summaries are factually inconsistent.

Our \textit{descriptions} differ from DeFacto \textit{explanations}: while both are expressed in natural language, the latter often conflate multiple errors within a single claim (see Table~\ref{tab:manual_proccessing_annotation}), which hinders the localization of each individual error. When considering the methodology for converting DeFacto's explanations to descriptions, we found that 
the conversion was often non-trivial and could not be automated. 
Moreover, we identified two major annotation problems: (1) many explanations contained issues such as missing or irrelevant information, vague phrasing, or incorrect annotations; 
and (2) a considerable number of factual inconsistencies were entirely missing from DeFacto.
While partial annotations can be useful for correction tasks, where any fix can improve the output, accurate \textit{evaluation} required a higher standard. We therefore designed a rigorous human annotation process, using DeFacto's annotations as a reference, to correct inaccurate labels and identify missing errors, ensuring our benchmark's reliability.
Since the annotation task is challenging we rely on expert annotators rather than crowdworkers: all annotations were performed by the authors of this paper. Given the scale of the dataset, we assign each example to 
a single annotator to balance quality with coverage.

\definecolor{colNone}{RGB}{235,223,202}
\definecolor{colExtr}{RGB}{111,192,217}
\definecolor{colDecomp}{RGB}{ 85,148,191}
\definecolor{colVague}{RGB}{224,224,224}
\definecolor{colMissing}{RGB}{96,96,96}
\definecolor{colRemove}{RGB}{104,176, 78}
\definecolor{colWrong}{RGB}{174, 61, 54}
\definecolor{colFabric}{RGB}{255, 99, 71} 

\begin{figure}[t]
  \centering
\resizebox{\linewidth}{!}{
\begin{tikzpicture}
\begin{axis}[
    width=20cm,
    height=2.5cm,        
    xbar stacked,
    ymin=-0.5, ymax=0.5,
    xmin=0, xmax=100,
    axis x line*=bottom,
    axis on top=true,
    xtick={0,20,40,60,80,100},
    xticklabel={\pgfmathprintnumber{\tick}\%},
    xticklabel style={font= \Large},
    tick style={black},
    tick align=outside,
    bar width=1,      
    ytick=\empty,     
    enlargelimits=false,
    legend columns=4,
    legend style={
    font=\fontsize{14pt}{16pt}\selectfont,
        /tikz/every even column/.append style={column sep=20pt},
        at={(0.5,2.5)}, anchor=north, draw=none,
    },
    legend cell align=left,
]

\addplot+[fill=colNone,draw=none, forget plot] coordinates {(24.6,0)};

\addplot+[fill=colExtr,draw=none, forget plot] coordinates {(48.7,0)};

\addplot+[fill=colDecomp,draw=none, forget plot] coordinates {(3.6,0)};

\addplot+[fill=colVague,draw=none, forget plot] coordinates {(8.1,0)};

\addplot+[fill=colMissing,draw=none, forget plot] coordinates {(2.6,0)};

\addplot+[fill=colRemove,draw=none, forget plot] coordinates {(5.6,0)};

\addplot+[fill=colWrong,draw=none, forget plot] coordinates {(3.6,0)};

\addplot+[fill=colFabric,draw=none, forget plot] coordinates {(3.5,0)};

\addlegendimage{area legend, fill=colNone, draw=colNone}
\addlegendimage{area legend, fill=colDecomp, draw=colDecomp}
\addlegendimage{area legend, fill=colMissing, draw=colMissing}
\addlegendimage{area legend, fill=colWrong, draw=colWrong}
\addlegendimage{area legend, fill=colExtr, draw=colExtr}
\addlegendimage{area legend, fill=colVague, draw=colVague}
\addlegendimage{area legend, fill=colRemove, draw=colRemove}
\addlegendimage{area legend, fill=colFabric, draw=colFabric}

\legend{
None,
Decomposition,
Missing Explanation,
Wrong Annotation,
Extraction,
Vague Explanation,
Irrelevant Information,
Fabricated}

\end{axis}
\draw[line width=0.5pt, draw=black] (rel axis cs:0,0) rectangle (rel axis cs:1,1);
\end{tikzpicture}
}
\caption{
    Estimated frequency of each annotation operation on the DeFacto explanations (see Table~\ref{tab:manual_proccessing_annotation}). \textit{None} refers to cases where the explanation described a single error and could be used as-is without modification. This estimation is based on 400 samples, details in \S\ref{Appendix:Curation Analysis}.}
\vspace{-5pt}
\label{fig:Operations distribution}
\end{figure}
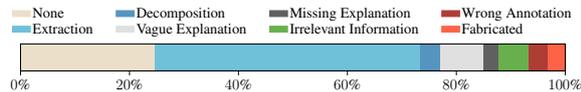

\paragraph{Phase 1: DeFacto Explanations to Descriptions.} 
Our benchmark is based on randomly selected 1,650 examples from DeFacto, consisting of 1,150 examples labeled as factually inconsistent and 500 labeled as factually consistent (Table~\ref{tab:dataset_stats}, first line). 
For each inconsistent example, we manually converted DeFacto's explanation into a list of descriptions. This process involved several operations: extracting and decomposing spans, revising vague descriptions, adding missing information, removing irrelevant content, and correcting inaccuracies. See Table~\ref{tab:manual_proccessing_annotation} for an example of each operation and \Cref{fig:Operations distribution} for their frequencies. 25\% of the explanations could be used without modification, 48\% required relatively straightforward error extraction, while the remaining 27\% involved \textit{challenging} annotation operations, that required expert annotators. 
Summaries classified as ``Fabricated'' were filtered-out, reducing the number of inconsistent examples from 1,150 to 1,086 (Table~\ref{tab:dataset_stats}, second line).
The annotation guidelines can be found in \S\ref{Appendix:DeFacto Curation}.

\paragraph{Phase 2: Error Enrichment via Human-LLM Collaboration}
During the annotation we observed that DeFacto's explanations often omit errors. This undermines our benchmark's reliability: a model that fails to detect a missing error will not get penalized, whereas a model that correctly identifies it will be unfairly penalized.

\begin{table}[t]
\centering
\large
\resizebox{\columnwidth}{!}{%
\begin{tabular}{lccc}
\toprule
\textbf{Source} & \textbf{Inconsistent} & \textbf{Consistent}& \textbf{Total} \\
\midrule
Original DeFacto  & 1,150 & 500 & 1,650\\
Phase 1: Explanations to Descriptions & 1,086 & 500 & 1,586\\
Phase 2: Error's Enrichment &  1,121 &  284 & 1,405 \\
\bottomrule
\end{tabular}%
}
\caption{Dataset statistics at each annotation phase.}
\vspace{-5pt}
\label{tab:dataset_stats}
\end{table}

Since detecting new errors is challenging, we perform LLM-assisted annotation \citep{ReframingHuman-AICollaboration,llmsformedicalannotation,unisumeval}, where candidate errors detected by an LLM are reviewed by human annotators. 
To increase coverage, we explicitly prompt the LLM to favor high recall, relying on the human annotators to filter-out false positives (more details in \S\ref{Appendix:Human-LLM Collaboration}).
This process increased the total amount of annotated errors from 1627 to 2131 (+31\%).
Importantly, out of the 500 summaries labeled as factually \textit{consistent} in DeFacto, 128 contained factual inconsistencies. 
In some cases the annotators couldn't determine whether the LLM-based suggestions were actual errors. To reduce subjectivity and maintain a reliable dataset, we filtered-out a total of 181 such examples. 
To validate this LLM-assisted annotation procedure, we double-annotated $150$ examples and found substantial inter-annotator agreement (raw agreement $=0.88$, Cohen's $\kappa=0.73$). Full details in \S\ref{Appendix:Human-LLM Collaboration}.

Final data statistics are presented in 
Table~\ref{tab:dataset_stats}, third line.
Figure~\ref{fig:Inconsistencies per Summary} presents a histogram of number of errors per-example. We randomly split the data into 140 development and 1,265 test examples.
\section{Experimental Setup}
\label{sec:exp_setup}
We use our benchmark to assess the capabilities of high-capacity LLMs on the task of fine-grained factual consistency evaluation.

\subsection{Models and Baselines}
\label{Baselines}

We evaluate GPT-4o-2024-11-20~\citep{gpt-4o}, Claude-3.5-sonnet-20241022~\citep{claude}, Gemini-1.5-pro~\citep{gemini-1.5-pro} and Llama-3.1-405B~\citep{llama}. Each evaluated model is prompted to perform the task end-to-end (\textbf{E2E}), namely to identify all inconsistencies and generate a list of descriptions to be passed to the LLM judge. We used \textbf{Zero-shot}, \mbox{\textbf{Few-shot}}, and Chain-of-Thought (\textbf{CoT}) prompting \citep{cot}. Full implementation details can be found in \S\ref{Appendix:E2E}. We compare to the following baselines:

\paragraph{Pipeline.} To compare to traditional evaluation pipelines, we implement a pipeline inspired by the \textbf{FactScore} approach \citep{factscore}. The evaluated LLM first decomposes the summary into atomic facts, and then assesses the factual consistency of each fact and, if any inconsistency is found, generates a description of it. Since a single fact may contain multiple inconsistencies, the model is instructed to describe each inconsistency individually.
This process can result in duplicate descriptions since multiple atomic facts may include the same erroneous information, and each fact is evaluated independently (see example in Figure~\ref{fig:factscore example}). To address this, the LLM is prompted to merge duplicate descriptions ensuring that each final description list contains one item for each inconsistency.\footnote{These steps are needed to evaluate error localization since the original implementation of FactScore only assesses each fact and returns a score that represent the fraction of consistent facts, which is not helpful for localizing specific errors.}

\begin{table*}[t]
\centering

\resizebox{\textwidth}{!}{
\small
\begin{tabular}{m{0.75cm} 
l 
c c c c  
c c c c 
c c c c  
c c c c 
c c c c  
}
\multirow{2}{*}{} & 
\multirow{2}{*}{} & 
\multicolumn{3}{c}{\normalsize GPT-4o} & 
& \multicolumn{3}{c}{\normalsize Claude-sonnet-3.5} & 
& \multicolumn{3}{c}{\normalsize Gemini-1.5-pro} & 
& \multicolumn{3}{c}{\normalsize Llama-3.1-405B} & &
& \multicolumn{3}{c}{\normalsize \textbf{Average}} \\
&& Rec. & Prec. & F1 & 
& Rec. & Prec. & F1 & 
& Rec. & Prec. & F1 & 
& Rec. & Prec. & F1 & 
&& Rec. & Prec. & F1 \\
\toprule
\multirow{3}{*}{E2E}  &Zero-Shot  & 0.35 & 0.70 & 0.47 & & 0.26 & \underline{\textbf{0.78}} & 0.40 & & 0.49 & 0.46 & 0.48 & & 0.53 & 0.50 & 0.51 & && 0.41 & 0.61 & 0.46 \\
&Few-Shot         & \underline{\textbf{0.56}} & 0.59 & 0.57 & & 0.44 & 0.74 & 0.56 & & 0.39 & 0.66 & 0.49 & & 0.48 & 0.57 & 0.52 & && 0.47 & 0.64 & 0.54 \\
&CoT              & 0.51 & 0.68 & \textbf{0.59} & & \underline{\textbf{0.67}} & 0.66 & \textbf{0.67} & & \underline{\textbf{0.54}} & 0.62 & 0.57 & & \textbf{0.54} & 0.59 & 0.56 & && \textbf{0.57} & 0.64 & \textbf{0.60} \\
\midrule
Pipeline & FactScore        & 0.52 & 0.66 & 0.58 & & 0.60 & 0.63 & 0.62 & & 0.30 & 0.69 & 0.42 & & 0.49 & 0.67 & \textbf{0.57} & && 0.48 & 0.66 & 0.55 \\
\midrule
\multirow{4}{*}{2-Step} &  Self + CoT & 0.41  &\textbf{0.76} & 0.54  & &0.51  & 0.76& 0.61 & & 0.39 &  \underline{\textbf{0.76}}& 0.52& & 0.38 &\underline{\textbf{0.76}}  &0.51 && &0.42  &\underline{\textbf{0.76}} &0.54 \\
&Self + CoT\&Hint & 0.42  &0.74 & 0.54& & 0.50& 0.77 & 0.61 & & 0.37 & 0.70 & 0.49&& 0.39 & 0.74  &0.51&& & 0.42 & 0.74 & 0.53\\
& TrueTeacher + CoT &0.47   & 0.74 &0.58  & & 0.62 &0.72  & \textbf{0.67} & & 0.49 & 0.70& \textbf{0.58} & &0.49  &0.66  &0.56 && &0.52& 0.71 &0.60  \\
& TrueTeacher + CoT\&Hint & 0.49 & 0.68 & 0.57 & & 0.61 & 0.72 & 0.66 & & 0.49 & 0.63 & 0.55 & & 0.49 & 0.60 & 0.54 && & 0.52 & 0.66 & 0.58 \\
\midrule
\multirow{1}{*}{2-Step} &Oracle + CoT  & 0.51 & \underline{0.77} &\underline{0.62}  & & \underline{0.67} &0.75 & \underline{0.71} & &\underline{0.54}  & 0.72 &\underline{0.61} & & 0.54 & 0.65 & \underline{0.59}&& & 0.57 &0.72  & \underline{0.63}\\
Oracle&Oracle + CoT\&Hint & 0.55 & 0.70 & \underline{0.62} & & \underline{0.67} & 0.75 & \underline{0.71} & & \underline{0.54} & 0.64 & 0.59 & & \underline{0.55} & 0.58 & 0.56 & && \underline{0.58} & 0.67 & 0.62 \\
\bottomrule
\end{tabular}
}
\caption{Performance of different LLMs on the \bench benchmark for \textbf{F}actual \textbf{I}nconsistencies \textbf{L}ocalization. In \textit{E2E}, the LLM performs end-to-end localization under various prompting strategies. In \textit{2-Step}, it localizes inconsistencies only when a preceding classifier flags the summary as inconsistent; in \textit{2-Step Oracle}, this is a perfect, oracle classifier. In \textit{Pipeline}, the LLM first decomposes the summary into atomic facts, which are then evaluated individually. Best non-oracle results per-column are in bold, best overall results are underlined. More details in \S \ref{Baselines}.}
\label{tab:llm_main_results}
\end{table*}

\paragraph{2-Step.} 
LLMs have demonstrated strong performance in \textit{binary} factual consistency evaluation \citep{trueteacher}, suggesting they can be used to improve \textit{fine-grained} evaluation by filtering-out cases which are unlikely to contain errors. Motivated by this, we explore two-step baselines: (Step 1) classify whether the summary is factually consistent; (Step 2) if classified as inconsistent, prompt the evaluated LLM to identify individual errors. For Step 1, we implement three variants: (1) \textbf{Self}, where the evaluated LLM performs binary classification using CoT prompting; (2) \textbf{TrueTeacher}, which uses the model from \citet{trueteacher};\footnote{\url{https://huggingface.co/google/t5_11b_trueteacher_and_anli}} and (3) \textbf{Oracle}, which uses the ground-truth label, serving as an upper bound.
For Step 2, we use \textbf{CoT} prompting, and also implement a \textbf{CoT\&Hint} variant, modifying the instruction to indicate that inconsistencies are present. Additional technical details are in \S\ref{Appendix:Baselines}.

\begin{figure}[t]
    \includegraphics[width=\columnwidth]{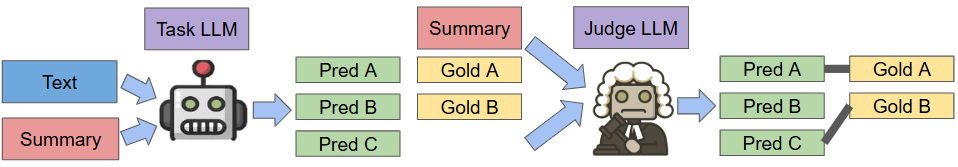}
    \caption{
    Illustration of our LLM-based evaluation protocol. The evaluated LLM detected three errors while the annotation listed two. The judge matches between the predicted and ground truth inconsistencies.}
    \label{fig:process description}
    \vspace{-5pt}
\end{figure}

\subsection{Evaluation}
\label{Experimental setup Evaluation}
We report error detection precision (the fraction predictions that are true inconsistencies), recall (the fraction of gold inconsistencies that are predicted), and F1.
To measure these metrics, we leverage LLM-as-a-judge as illustrated in Figure~\ref{fig:process description}. The judge receives the summary, the list of gold descriptions, and a list of predicted descriptions, and is prompted with a \textit{matching} task: aligning items from the gold list with those in the predicted list. Matched descriptions are counted as true positives. We use GPT-4o as the judge. To validate the judgment's quality, we conduct extensive human evaluation on our 140-sample development set, generating outputs for each prompt by each model. We assign human annotators with the same matching task as the judge model to create human-annotated matches, to serve as ground truth.
We then calculate the \textit{precision} and \textit{recall} of the matching task, which were 0.95 and 0.92, respectively, providing evidence that our approach produces high-quality judgments. Additional implementation details on the judgment process can be found in \S\ref{Appendix:Evaluation}.

\section{Results}
Table~\ref{tab:llm_main_results} presents the main results. 
The cross-model average F1 (rightmost column) remains below 0.60 for all methods (except with the Oracle classifier), highlighting the benchmark's difficulty.

In the E2E setup, CoT shows superior performance, suggesting that reasoning is helpful for localizing factual inconsistencies. 
Interestingly, CoT surpasses FactScore (average F1 of 0.60 vs. 0.55), suggesting that allowing the model to reason freely is more effective than controlling its reasoning process through predefined steps. Another notable trend is that precision consistently exceeds recall, suggesting that LLMs tend to focus on a subset of errors for which they have sufficient confidence.

We next analyze the effectiveness of the preliminary filtering step in the 2-Step setting. \textit{Oracle+CoT} filters error-free examples, thereby improving precision and outperforming \textit{CoT}.
Conversely, \textit{Self+CoT} underperforms \textit{CoT}, while \textit{TrueTeacher+CoT} only matches \textit{CoT}'s performance.
These results highlight the potential of perfect filtering to improve performance, but also suggest that with current classification quality, end-to-end approaches remain more effective.

\begin{figure}
  \centering
\resizebox{\linewidth}{!}{
\begin{tikzpicture}[trim left=-5pt, baseline]

\begin{groupplot}[%
   group style={group size=4 by 1, horizontal sep=0.3cm},
  width=6cm, height=1.5cm,
  ybar,
  x=0.7cm,
  enlarge x limits=0.25,
  axis line style={very thin},
  ymin=0, ymax=1.1,
  scale only axis,
  symbolic x coords={F1, Prec, Rec.},
  xtick=data,
  axis x line*=bottom,
  axis y line*=left,
  tick style={very thin},
  label style={font=\scriptsize},
  tick label style={font=\scriptsize},
  title style={font=\scriptsize},
  axis on top,
  legend style={at={(-1.25,-0.375)}, anchor=north, legend columns=-1, font=\scriptsize},
]

\nextgroupplot[title=\small GPT-4o,title style={yshift=-5pt},ytick={0.2,0.4,0.6,0.8,1.0},
  yticklabels={0.2, 0.4, 0.6, 0.8, 1.0},tick align=outside,]
  \addplot+[bar width=7pt, bar shift=-3.2pt,fill=blue!70,draw=blue!70] coordinates
    {(F1,0.88) (Prec,0.87) (Rec.,0.88)};
  \addplot+[bar width=7pt,fill=cyan!30!white,draw=cyan!30!white]  coordinates
    {(F1,0.75) (Prec,0.94) (Rec.,0.63)};

\nextgroupplot[title=\small Llama 3.1 405B, title style={yshift=-5pt},yticklabels={},
  ytick=\empty]
  \addplot+[bar width=7pt,bar shift=-3.2pt,fill=blue!70,draw=blue!70] coordinates
    {(F1,0.92) (Prec,0.86) (Rec.,0.98)};
  \addplot+[bar width=7pt,fill=cyan!30!white,draw=cyan!30!white]  coordinates
    {(F1,0.73) (Prec,0.97) (Rec.,0.59)};

\nextgroupplot[title=\small Claude sonnet 3.5, title style={yshift=-5pt},yticklabels={},
  ytick=\empty]
  \addplot+[bar width=7pt,bar shift=-3.2pt,fill=blue!70,draw=blue!70] coordinates
    {(F1,0.89) (Prec,0.83) (Rec.,0.96)};
  \addplot+[bar width=7pt,fill=cyan!30!white,draw=cyan!30!white]  coordinates
    {(F1,0.77) (Prec,0.96) (Rec.,0.65)};

\nextgroupplot[title=\small Gemini 1.5 pro, title style={yshift=-7pt},yticklabels={},
  ytick=\empty]
  \addplot+[bar width=7pt,bar shift=-3.2pt,fill=blue!70,draw=blue!70] coordinates
    {(F1,0.86) (Prec,0.83) (Rec.,0.9)};
  \addplot+[bar width=7pt,fill=cyan!30!white,draw=cyan!30!white]  coordinates
    {(F1,0.7) (Prec,0.95) (Rec.,0.56)};

\legend{Binarized, Binary}
\end{groupplot}
\end{tikzpicture}
}

\caption{Performance on \textit{binary} factual consistency evaluation. In ``Binary'' the LLM is prompted with the binary task, while in ``Binarized'' it is prompted for fine-grained evaluation and its outputs are post-processed into binary labels.}

\vspace{-5pt}
\label{fig:binary_to_fine_comparison}
\end{figure}
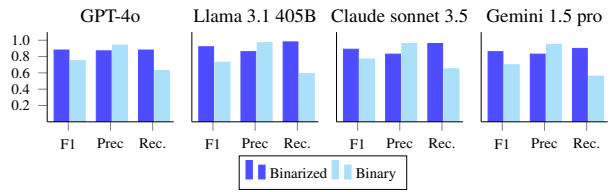

The fact that \textit{Self+CoT}, where the same LLM first performs a filtering step, underperforms \textit{CoT} is rather surprising, since LLMs are expected to perform well in the binary classification task. 
To further understand this gap, we focus on the \textit{binary} factual consistency evaluation task and compare each LLM's performance against a \textit{Binarized} baseline, where the LLM is prompted to perform \textit{fine-grained} evaluation followed by post-processing that labels a summary as inconsistent if at least one error is detected.\footnote{We use the CoT fine-grained variant in \textit{Binarized} and prompt the model to reason step-by-step in \textit{Binary}.} The results are presented in Figure~\ref{fig:binary_to_fine_comparison}. \textit{Binarized} achieves higher F1, despite \textit{Binary} assigning the model a seemingly easier task. One possible explanation is that in \textit{Binarized}, the model must identify all inconsistencies, which may encourage a more thorough analysis of the content.
Interestingly, \textit{Binary} consistently yields higher precision but lower recall compared to \textit{Binarized}. This suggests that the model is more conservative in the \textit{Binary} setup, avoiding false positives but failing to detect many actual inconsistencies. 

Lastly, we examine the effect of explicitly informing the model that the summary contains errors by comparing \textit{Oracle+CoT} to \textit{Oracle+CoT\&Hint}. Since these variants share the same filtering step, they allow us to directly compare \textit{CoT} to \textit{CoT\&Hint} on the same examples. 
As expected, \textit{CoT\&Hint} achieves higher recall than \textit{CoT}, as the hint encourages the model to identify more errors. However, this comes at a significant cost to precision, suggesting the model becomes overly permissive and flags many false positives, ultimately reducing overall F1.

\begin{table*}[!ht]
\centering
\resizebox{\textwidth}{!}{
\begin{tabular}{>{\arraybackslash}p{1.75cm} >{\arraybackslash}p{4.9cm} p{6cm} p{4.6cm} >{\arraybackslash}p{4cm}}
\hline
\multicolumn{1}{c}{\textbf{Category}} & 
\multicolumn{1}{c}{\textbf{Definition}} &
\multicolumn{1}{c}{\textbf{Summary}} & 
\multicolumn{1}{c}{\textbf{Text Evidence}} & 
\multicolumn{1}{c}{\textbf{Explanation}}\\
\hline
\multirow{2}{=}{\centering Extrinsic\\Correct} &\small  The inconsistency is additional information in the summary that is not in the source text, and is factually correct. & \small Japan's Hayabusa2 spacecraft landed on \textcolor{red}{Ryugu}, collected samples, and has returned them to Earth for solar system research. & \small A Japanese spacecraft successfully landed on \textcolor{red}{an asteroid}... & \small The name of the astroid "Ryugu" is correct, but does not appear in the text \\[4ex] 
\hline
\multirow{2}{=}{\centering  Extrinsic\\Wrong}  & \small The inconsistency is additional information in the summary that is not in the source text and is factually incorrect. & \small Japan's Hayabusa2 spacecraft landed on \textcolor{red}{Bennu}, collected samples, and has returned them to Earth for solar system research.  & \small A Japanese spacecraft successfully landed on \textcolor{red}{an asteroid}... & \small The name of the astroid "Bennu" is wrong, and it does not appear in the text  \\[4ex] 
\hline
\multirow{2}{=}{\centering Intrinsic\\Alteration} &\small The inconsistency is based on information from the source text that has been altered in the summary. & \small Japan's Hayabusa2 spacecraft landed on an asteroid, collected samples, and \textcolor{red}{ will return} them to Earth for solar system research. & \small ...and \textcolor{red}{later returned} the samples to Earth... & \small The summary claims it "will return", but according to the text it returned. \\[4ex] 
\hline
\multirow{2}{=}{\centering Intrinsic\\Composition} &  \small The inconsistency is the result of individually correct facts from the source text being combined poorly. & \small Japan's Hayabusa2 spacecraft landed on an asteroid, collected samples, and has returned them to Earth for \textcolor{red}{groundbreaking}  solar system research.& \small  ...collected samples in a \textcolor{red}{groundbreaking mission}...to \textcolor{red}{learn more about the origins of the solar system}... & \small The summary calls the research "groundbreaking," but that's the description of the mission.\\[4ex] 
\hline
\end{tabular}
}
\captionsetup{
  skip=0pt,
  belowskip=0pt
}
\caption{Definitions and examples of false negatives, cases where the model failed to detect factual inconsistencies, accompanied by an explanation to why each example was classified to that category.
}
\label{tab:Examples of categories in the false negatives analysis}
\end{table*}

\section{Error analysis}
\label{error analysis}
This section presents an error analysis to better understand the reasons models make mistakes.
We divide it into 2 parts: (1) \textbf{false negatives analysis} for why models do not detect some inconsistencies, and (2) \textbf{false positives analysis} for why some of the models' predictions are incorrect.

\begin{figure}
\centering
\resizebox{\linewidth}{!}{%
\begin{tikzpicture}
\begin{axis}[
  width=10cm,
  height=3.5cm,
  colormap/YlGnBu,
  colorbar,
  colorbar style={ width=0.25cm, ytick={0,20,40,60,80,100},
     yticklabels={0\%,20\%,40\%,60\%,80\%,100\%},
     tick label style={font=\footnotesize},
     tick align=outside,ytick pos=right},
  xmin= -0.5, xmax=4.5, 
  ymin= -0.5, ymax=3.5,
  enlargelimits=false,
  xtick=data,
  ytick=data,
  yticklabels={
  {\parbox{2.8cm}{\centering Extrinsic Correct}},
  {\parbox{2.8cm}{\centering Extrinsic Wrong}},
  {\parbox{2.8cm}{\centering Intrinsic Alteration}},
  {\parbox{2.8cm}{\centering Intrinsic Composition}}
},
yticklabel style={align=center,text width=2.8cm},
  yticklabel style={align=center,font=\small},
  xticklabels={{
    Llama\\3.1-405b},
    {GPT 4o},
    {Claude\\3.5-sonnet},
    {Gemini\\1.5-pro},
    {Average}},
  xticklabel style={align=center,font=\small},
  y dir=reverse,
 axis lines=box,
xtick pos=bottom,
ytick pos=left,
tick style={draw=black, thick},
tick align=outside,
]

\pgfplotstableread[row sep=crcr]{%
x y z\\
0 0 62\\ 1 0 68\\ 2 0 71\\ 3 0 59\\ 4 0 65\\
0 1 11\\ 1 1 11\\ 2 1  5\\ 3 1 19\\ 4 1 12\\
0 2 21\\ 1 2 15\\ 2 2 20\\ 3 2 20\\ 4 2 19\\
0 3  6\\ 1 3  5\\ 2 3  4\\ 3 3  3\\ 4 3 4\\
}\heatmap

\pgfplotstablecreatecol[
  create col/assign/.code={
    \pgfmathtruncatemacro\res{\thisrow{z}>50}
    \pgfkeyslet{/pgfplots/table/create col/next content}\res
  }
]{greaterthan50}{\heatmap}

\addplot [
  matrix plot*,
  mesh/rows=4, mesh/cols=5,
  point meta=explicit,
] table[row sep=crcr, meta=z]{\heatmap};

\addplot [
  scatter,
  only marks,
  mark=none,
  point meta=explicit,
  x filter/.code={
    \pgfplotstablegetelem{\coordindex}{[index]3}\of\heatmap
    \ifnum\pgfplotsretval=1
    \else
      \def\pgfmathresult{inf}
    \fi
  },
  unbounded coords=discard,
  nodes near coords={\pgfmathprintnumber[precision=0]{\pgfplotspointmeta}\%},
  every node near coord/.append style={
    font=\small,
    anchor=center,
    text=white,
  },
] table[x=x,y=y,meta=z]{\heatmap};

\addplot [
  scatter,
  only marks,
  mark=none,
  point meta=explicit,
  x filter/.code={
    \pgfplotstablegetelem{\coordindex}{[index]3}\of\heatmap
    \ifnum\pgfplotsretval=0
    \else
      \def\pgfmathresult{inf}
    \fi
  },
  unbounded coords=discard,
  nodes near coords={\pgfmathprintnumber[precision=0]{\pgfplotspointmeta}\%},
  every node near coord/.append style={
    font=\small,
    anchor=center,
    text=black,
  },
] table[x=x,y=y,meta=z]{\heatmap};

\end{axis}
\end{tikzpicture}%
}
\caption{Distribution of false negatives (Table~\ref{tab:Examples of categories in the false negatives analysis}).}
\label{fig:heatmap_fn}
\vspace{-10pt}
\end{figure}
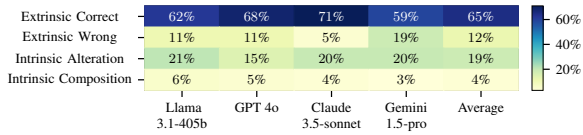

\paragraph{False Negatives.}
We manually analyzed a random sample of 150 undetected inconsistencies from each model and categorized them into four categories. We present their definitions and examples in Table~\ref{tab:Examples of categories in the false negatives analysis}, and 
their distribution in Figure~\ref{fig:heatmap_fn}. 
The most common category shared across all models is \textit{Extrinsic Correct}. Since the errors in this category involve factually correct information, we hypothesize that a key reason for these failures is that the information aligns with the model's parametric knowledge, making it difficult for the model to recognize it as inconsistent, even when it is not supported by the source text. To substantiate this hypothesis, we analyze (1) whether the information in question indeed aligns with the model's knowledge, and (2) whether this alignment is the reason the model fails to identify these inconsistencies.

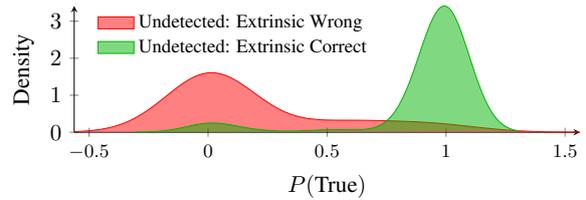
\begin{figure}
\centering
\resizebox{\linewidth}{!}{%
\begin{tikzpicture}
\begin{axis}[
xlabel style={font=\small},
ylabel style={font=\small},
    xlabel={$P(\text{True})$},
    ylabel={Density},
    width=10cm,
    height=3.7cm,
    axis lines=left,
    ymin=0,
    clip=true,
    every axis plot/.append style={fill opacity=0.5},
    cycle list name=color list,
    legend style={
        at={(0.6,1)},
        anchor=north east,
        draw=none,
        fill=white,
        font=\small  ,
        legend columns=1,
         legend cell align={left}
    },
    xlabel near ticks,
    ylabel near ticks,
    xticklabel style={align=center,font=\small},
]

\addplot+[red, mark=none, fill=red, forget plot]
    table [x=x, y=y, col sep=comma] {figs_full_latex/p_true_extrinsic_wrong_kde.csv};
\addlegendimage{area legend, fill=red, draw=red}
\addlegendentry{Undetected: Extrinsic Wrong}

\addplot+[green!70!black, mark=none, fill=green!70!black, forget plot]
    table [x=x, y=y, col sep=comma] {figs_full_latex/p_true_extrinsic_correct_kde.csv};
\addlegendimage{area legend, fill=green!70!black, draw=green!70!black}
\addlegendentry{Undetected: Extrinsic Correct}

\end{axis}
\end{tikzpicture}
}

\caption{Density of $\mathbf{P(True)}$ scores for the \textit{Extrinsic Correct} and \textit{Extrinsic Wrong} false negatives (Table~\ref{tab:Examples of categories in the false negatives analysis}).}
\label{fig:p_true_density}
\vspace{-5pt}
\end{figure}

To provide evidence that the \textit{Extrinsic Correct} errors contain information that is mostly known to the model we need a method to assess LLMs' knowledge, which is not trivial \citep{fierro-etal-2024-defining, gekhman-etal-2024-fine, insideout}. We choose to use $\mathbf{P(True)}$ \citep{p_true}, a popular metric that quantifies the likelihood that the model assigns for the correctness of a specific answer to a question. For each inconsistency, we ask human annotators to generate a question $\mathbf{q}$ for which the answer $\mathbf{a}$ is the factually inconsistent information from the summary.\footnote{
Technical details are in \S\ref{Appendix: False Negative Analysis} and an example is presented in Figure~\ref{fig:armstrong-example}. We ran this analysis for GPT-4o.} We then calculate $\mathbf{P(True\mid q, a)}$, as an estimate to whether the factually inconsistent information aligns with the model's knowledge. 
Figure~\ref{fig:p_true_density} presents a density plot of $\mathbf{P(True)}$ scores. For the \textit{Extrinsic Correct} category, the distribution is concentrated near 1, indicating that it contains facts that are largely known to the model.

After establishing that the model often possesses knowledge about the correct information added in \textit{Extrinsic Correct} cases, we provide evidence that this may explain its failure to detect such errors. We ask human annotators to generate counterfactual versions for \textit{Extrinsic Correct} inconsistencies by replacing the (correct) added information with semantically similar but incorrect alternatives. For example, if a summary says \textit{``the protests were in London''}, and \textit{``London''} is \textit{not} in the source text but is correct, we might replace it with a different UK city. We found that 88.1\% of these counterfactual errors were successfully detected. This result, together with the P(True) analysis, strongly suggest that the alignment with the model's parametric knowledge causes it to miss that the added information is unsupported by the source text.

\paragraph{False Positives.}
We manually analyzed a random sample of 100 false positives per-model: predictions that do not reflect real factual inconsistencies.
We have identified 5 main categories, with an example of each presented in Table \ref{tab:false-postives-categories} in the Appendix.

\vspace{-5pt}
\begin{itemize}
  \item \textbf{Overlooked Info.} Failure to recognize information that is explicitly stated in the source text, leading to wrong prediction.
  
  \vspace{-5pt}
  \item \textbf{Missed Deduction.} Failure to recognize a fact that can be directly deduced from the text.
  
  \vspace{-5pt}
  \item \textbf{Omission.} Classifying information that is missing in the summary as an inconsistency.
  \vspace{-5pt}
  \item \textbf{Overly literal.} Classifying superficial changes in wording as an inconsistency.
  \vspace{-5pt}
  \item \textbf{Invented.} The information that is mentioned as inconsistent is not in the text or summary.
\end{itemize}
\vspace{-5pt}

\begin{figure}[!t]
\centering
\resizebox{\linewidth}{!}{%
\begin{tikzpicture}
\begin{axis}[
  width=10cm,
  height=4cm,
  colormap/YlGnBu,
  colorbar,
  colorbar style={width=0.2cm,ytick={0,10,20,30,40,50,60,70,80,90,100},
    yticklabels={0\%,10\%,20\%,30\%,40\%,50\%,60\%,70\%,80\%,90\%,100\%},
    tick label style={font=\footnotesize},
    tick align=outside,ytick pos=right},
  xmin= -0.5, xmax=4.5, 
  ymin= -0.5, ymax=4.5,
  enlargelimits=false,
  xtick=data,
  ytick=data,
 yticklabels={
  {\parbox{2.35cm}{\centering Overlooked Info}},
  {\parbox{2.35cm}{\centering Missed Deduction}},
  {\parbox{2.35cm}{\centering Omission}},
  {\parbox{2.35cm}{\centering Overly Literal}},
  {\parbox{2.35cm}{\centering Invented}}
},
yticklabel style={align=center,text width=2.35cm,font=\small},
  yticklabel style={align=center},
  xticklabels={{
    Llama\\3.1-405b},
    {GPT 4o},
    {Claude\\3.5-sonnet},
    {Gemini\\1.5-pro},
    {Average}}, 
  xticklabel style={align=center,font=\small},
  y dir=reverse,
 axis lines=box,
xtick pos=bottom,
ytick pos=left,
tick style={draw=black, thick},
tick align=outside,
]

\pgfplotstableread[row sep=crcr]{%
x y z\\
0 0 12\\ 1 0 13\\ 2 0 23\\ 3 0 32\\ 4 0 20\\
0 1 42\\ 1 1 31\\ 2 1  49\\ 3 1 13\\ 4 1 34\\
0 2 34\\ 1 2 41\\ 2 2 14\\ 3 2 36\\ 4 2 31\\
0 3  10\\ 1 3  11\\ 2 3  12\\ 3 3  8\\ 4 3 10\\
0 4  2\\ 1 4  4\\ 2 4  2\\ 3 4  11\\ 4 4 5\\
}\heatmap

\pgfplotstablecreatecol[
  create col/assign/.code={
    \pgfmathtruncatemacro\res{\thisrow{z}>25}
    \pgfkeyslet{/pgfplots/table/create col/next content}\res
  }
]{greaterthan25}{\heatmap}

\addplot [
  matrix plot*,
  mesh/rows=5, mesh/cols=5, 
  point meta=explicit,
] table[row sep=crcr, meta=z]{\heatmap};

\addplot [
  scatter,
  only marks,
  mark=none,
  point meta=explicit,
  x filter/.code={
    \pgfplotstablegetelem{\coordindex}{[index]3}\of\heatmap
    \ifnum\pgfplotsretval=1
    \else
      \def\pgfmathresult{inf}
    \fi
  },
  unbounded coords=discard,
  nodes near coords={\pgfmathprintnumber[precision=0]{\pgfplotspointmeta}\%},
  every node near coord/.append style={
   font=\small,
    anchor=center,
    text=white,
  },
] table[x=x,y=y,meta=z]{\heatmap};

\addplot [
  scatter,
  only marks,
  mark=none,
  point meta=explicit,
  x filter/.code={
    \pgfplotstablegetelem{\coordindex}{[index]3}\of\heatmap
    \ifnum\pgfplotsretval=0
    \else
      \def\pgfmathresult{inf}
    \fi
  },
  unbounded coords=discard,
  nodes near coords={\pgfmathprintnumber[precision=0]{\pgfplotspointmeta}\%},
  every node near coord/.append style={
    font=\small,
    anchor=center,
    text=black,
  },
] table[x=x,y=y,meta=z]{\heatmap};

\end{axis}
\end{tikzpicture}%
}
\caption{Distribution of False Positive categories.}
\label{fig:heatmap_fp}
\vspace{-5pt}
\end{figure}
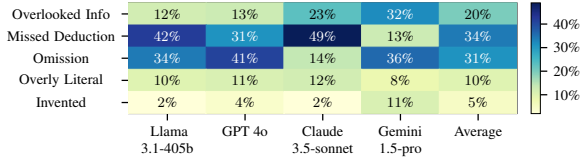

As shown in Figure \ref{fig:heatmap_fp}, the most prevalent categories were \textit{Missed Deduction} and \textit{Omission}. The former is expected, as identifying information that is not explicitly stated in the text is challenging. The latter is rather surprising, as it suggests limited instruction following capabilities. The prompt clearly instructs to identify facts \textit{in the summary} that cannot be verified, yet the model flags information that is simply omitted from the summary.

\section{Related Work}
\label{sec:related}
Previous work on factual consistency evaluation can be categorized along two axes: (1) the setting is either \textit{binary}, assigning a single score to the entire output, or \textit{fine-grained}, localizing specific errors; and (2) the goal is either \textit{evaluation}, focusing on building consistency-checking systems, or \textit{meta-evaluation}, focusing on evaluation of such systems.

Most work in the \textit{fine-grained} setting focus on \textit{evaluation} and not \textit{meta-evaluation}.
In these studies, factual inconsistencies are represented as entities \citep{cao2022}, sentences \citep{Summac}, spans \citep{XSumFaith,cliff,ragtruth,Colm_fine_grain_detection_and_edit}, atomic facts \citep{factscore,propsegment,wice} or QA pairs \citep{q2, wang2020asking, qafacteval, cattan2024localizing}. As we discuss in \S\ref{sec:representation}, the expressivity of these representations is limited, constraining the range of errors that can be captured, and they can often be vague, which complicates evaluation. 

Perhaps owing to the evaluation challenges stemming from existing error representations, most work on \textit{meta-evaluation} of factual consistency focus on the \textit{binary} setting \citep{True,Summac,trueteacher,seahorse,AGGREFACT,luo2023chatgpt}. 
To our knowledge, the only work that performed meta-evaluation in the fine-grained setting is \citet{ragtruth}. They represent errors using spans and measure character-level overlap, a limitation we discuss in detail in \S\ref{sec:representation}. In addition, their evaluation focused on 
GPT-4-turbo \cite{openai2023gpt4}, GPT-3.5-turbo \cite{openai2022chatgpt}, and Llama-2-13B \cite{touvron2023llama}, leaving the performance of the highest-capacity models as an open question.

Our work addresses the limitations of existing error representations by proposing a new one based on textual descriptions. Not only does it allow us to capture the full range of possible errors, but when combined with our LLM-based evaluation protocol, it also helps us overcome the evaluation challenge of comparing predicted factual inconsistencies to ground truth ones. This framework facilitates the construction of \bench, a high-quality benchmark for the meta-evaluation of LLMs on the task. In addition, our study evaluates extremely high-capacity LLMs, providing a fresh perspective on their capabilities at a scale not previously studied for this task. Finally, our detailed error analysis sheds light on the reasons for their failures.

\section{Conclusion}

We take a step towards replacing existing fine-grained factual consistency evaluation systems with LLMs. 
We introduce \bench \xspace - the first benchmark for \textbf{F}actual \textbf{I}nconsistencies \textbf{L}ocalization using LLMs, with 1,400 carefully annotated examples. We evaluate four strong LLMs, with a detailed analysis that offers a clear view of their strengths and weaknesses. We hope that our benchmark and insights will foster a shift towards LLM-based evaluation, which will support broader adoption of fine-grained consistency evaluation in practical, real-world applications.

\clearpage

\section{Limitations}

\paragraph{Exclusive Focus on LLMs.}
A key limitation of our work is that the proposed benchmark and evaluation protocol are specifically tailored for LLMs. We introduce an error representation based on free-form textual descriptions, which aligns naturally with the capabilities of LLMs and, when paired with our LLM-based evaluation protocol, resolves evaluation challenges caused by the vagueness of prior representations. However, this design choice makes it difficult to use our framework to directly compare LLMs with traditional, non-LLM systems that rely on different output formats, such as entities, spans, or QA pairs.

We made a considerable effort to bridge this gap with our \textbf{Pipeline} baseline (see \S\ref{Baselines}), which emulates a traditional evaluation method by \cite{factscore}.
However, as can be seen in \textbf{Pipeline}'s implementation details, adapting \citeauthor{factscore} method for a direct comparison proved to be a non-trivial task, requiring substantial modifications to make its output compatible with our description-based framework, highlighting the inherent difficulty of such cross-paradigm comparisons.

In this context, we would like to highlight that the choice of error representation typically fundamentally impacts not only the system design but also the ability to annotate gold labels and create a standardized benchmark, since comparing systems with disparate output formats is a significant challenge. Thus, our work deliberately focuses on a representation suitable for LLMs to establish a high-quality benchmark for their meta-evaluation, acknowledging that this specialization limits its applicability for evaluating systems with different architectures.

\paragraph{Benchmark Annotation Challenges and Coverage.}
Creating a high-quality, comprehensive benchmark for fine-grained hallucination detection is an extremely challenging task. The difficulty of achieving complete annotation coverage is likely inherent to any benchmark for such a complex and subjective task. This was evident in the original DeFacto dataset, where we found numerous issues, including vague or incorrect annotations and a considerable number of factual inconsistencies entirely missing from the dataset. To ensure the reliability of our \bench benchmark, we took several quality control steps, such as relying on expert annotators and implementing a rigorous, two-phase annotation process that included an LLM-human collaboration to enrich the data, which increased the amount of discovered inconsistencies by 31\% (see \S\ref{Benchmark}).

Despite these extensive measures, we acknowledge that some inconsistencies may still be unannotated. While this is a common challenge for such datasets, we believe in being transparent about it.

\bibliography{refernces}

\appendix

\section{Dataset}
\subsection{DeFacto Curation}
\label{Appendix:DeFacto Curation}
As noted in \S\ref{Benchmark}, our benchmark builds on the DeFacto Benchmark \citep{DeFacto}, which studied how human feedback improves models at correcting factual inconsistencies. DeFacto paired texts with summaries and, when inconsistencies were found, provided feedback to correct the summary. This feedback included three main components: (1) an \textbf{explanation} of the inconsistencies in the summary, (2) \textbf{instructions} of how to correct the summary and (3) a \textbf{revised summary} fixed by a human. Ideally, we could simply use the explanations provided in the dataset, but a considerable portion of the original explanations were lacking. Beyond the fact that numerous inconsistencies were missing (more on that in \S\ref{Appendix:Human-LLM Collaboration}), the provided explanations were often not usable as is: some inconsistencies were merged together, some explanations were vague or incomplete, and others contained irrelevant information. Therefore, we extracted individual inconsistencies descriptions from DeFacto through manual annotation.
Participants were instructed to review the annotations provided in the original DeFacto dataset and extract descriptions of factual inconsistencies. Then they were asked to rephrase them as individual natural language statements explicitly identifying the inconsistency in the summary, with minimal changes to the original dataset.

Below are the annotation instructions:\\\\
\newtcolorbox{guidelinesbox}{
  colback=green!10!white,    
  colframe=black,     
  boxrule=0.5pt,      
  arc=0pt,            
  outer arc=0pt,
  left=5pt, right=5pt, top=5pt, bottom=5pt, 
  fonttitle=\bfseries,
  coltitle=black,
  breakable,
  fontupper=\small\fontfamily{phv}\selectfont
}
\begin{guidelinesbox}
\textbf{\large Task Overview}\\\\
You will be provided with the following for each sample:
\begin{itemize}[left=0pt, itemsep=0pt]
    \item Text: The source document.
    \item Summary: A factually inconsistent summary of the text.
    \item Raw Human Annotation, which includes:
    \begin{itemize}[left=0pt, itemsep=0pt, topsep=0pt]
        \item A written explanation of the inconsistencies in the summary.
        \item Instructions for fixing the summary.
        \item An edited summary that is factually consistent.
    \end{itemize}
\end{itemize}
Your task is to extract individual \textbf{factual inconsistency descriptions}—short, self-contained statements that identify exactly what information in the summary is factually inconsistent with the source text.\\\\
\textbf{\large Instructions}
\begin{itemize}[left=0pt, itemsep=0pt]
    \item \textbf{Extract Individual Inconsistencies}
    \begin{itemize}
        \item For each factual inconsistency described in the explanation, extract and formulate it as a complete, standalone sentence that clearly identifies the inconsistency on its own, without referring to other inconsistencies.
        \item Each description should explicitly state what is incorrect in the summary.
    \end{itemize}
    \item \textbf{Minimize Changes}
    \begin{itemize}[left=0pt, itemsep=0pt, topsep=0pt]
        \item Make minimal edits to the original wording of the explanation.
        \item Modify the explanation only when necessary:
        \begin{itemize}[left=0pt, itemsep=0pt, topsep=0pt]
            \item To phrase it as a standalone sentence.
            \item If the inconsistency is vague or unclear → rephrase it to be precise and unambiguous.
            \item The explanation includes irrelevant or redundant information → remove any parts that do not describe factual inconsistencies (e.g., mentions of consistent facts, mentions of information that is omitted from the summary, context and so on).
        \end{itemize}
    \end{itemize}
    \item \textbf{Add Missing Inconsistencies}\\
    If an inconsistency is evident in the edited summary or instructions but is not mentioned in the explanation, write a new description for it.
    \item \textbf{Remove Invalid Annotations}\\
    If you notice that a described inconsistency is not actually inconsistent—for example, the information is not present in the summary, or the summary is consistent with the text—discard it.
    \item \textbf{Remove Low-Quality Samples}\\
    If the entire summary is factually inconsistent (e.g., instructions call for a complete rewrite), mark the sample as "discarded: full rewrite required" and do not extract any descriptions.
\end{itemize}
\end{guidelinesbox}
\subsection{Curation Analysis}
\label{Appendix:Curation Analysis}
After converting the original DeFacto dataset explanations into our description based format, as outlined in \S\ref{Benchmark} and \S\ref{Appendix:DeFacto Curation}, in order to understand the extent of changes to the original dataset, we performed the following analysis: We sampled 400 text-summary pairs from the DeFacto dataset, all originally labeled as factually inconsistent. These samples contained a total of 655 identified inconsistencies. We manually analyzed the required operations to adapt the original DeFacto explanations into our format. We categorized the operations into two main types, further divided into eight subcategories:
\begin{itemize}[left=0pt, itemsep=0pt]
    \item \textbf{Simple Changes } - These included cases where the explanation required no modification at all, or where inconsistencies were directly extracted from the explanation.
    \item \textbf{Challenging Changes} - These involved more significant intervention, including decomposition of merged inconsistencies, removal of irrelevant information, clarification of vague explanations, inferring missing inconsistencies not originally annotated, identifying incorrect annotations, and discarding summaries that were almost entirely unrelated to the source text.
\end{itemize}
Table \ref{tab:manual_proccessing_annotation} provides a detailed illustration of the various operations, accompanied by examples.
Out of the 655 inconsistencies, the distribution of required operations was as follows:
\begin{itemize}[left=0pt,itemsep=0pt]
    \item \textbf{Simple Changes: 480 (73.3\%)}
    \begin{itemize}[itemsep=0pt,left=1pt]
        \item No change needed: 161 (24.6\%)
        \item Extraction: 319 (48.7\%)
    \end{itemize}
    \item \textbf{Challenging Changes: 175 (26.7\%)}
    \begin{itemize}[itemsep=0pt,left=1pt]
        \item Decomposition: 24 (3.6\%)
        \item Clarifying vague explanation: 53 (8.1\%)
        \item Imputing missing explanation: 17 (2.6\%)
        \item Removal of irrelevant information: 37 (5.6\%)
        \item Removal of incorrect annotation: 21 (3.2\%)
        \item Removal of unrelated summary: 23 (3.5\%)
    \end{itemize}
\end{itemize}

\subsection{Human-LLM Collaboration}
\label{Appendix:Human-LLM Collaboration}
During the process of converting DeFacto explanations into descriptions (as detailed in \S\ref{Benchmark} and \S\ref{Appendix:DeFacto Curation}), and through initial experiments using an LLM as annotator, it became clear that many inconsistencies were not present in the original dataset. At the same time, initial experiments revealed that a significant number of inconsistencies surfaced by LLMs were valid, even though they did not appear in the original dataset. Since the DeFacto dataset was annotated by humans, we hypothesized that it was less likely that further human annotation would uncover many new inconsistencies, and using LLMs blindly would have introduced a lot of false inconsistencies into our dataset. Therefore, we decided to use a collaborative approach, leveraging the LLM's ability to surface new, unannotated inconsistencies with human judgment. The approach was based on high recall prompting and human filtering.

\paragraph{High Recall Prompting and Human Filtering.}
An LLM is prompted to identify as many potential inconsistencies as possible, explicitly prioritizing high recall over precision. We used GPT-4o for this purpose and provide the full prompt in \S\ref{Appendix:High Recall Prompt}. Human annotators then reviewed the LLM's outputs to filter out false positives. This process yielded additional true inconsistencies that were originally missed by human annotators. Using this method, we managed to increase the amount of annotated inconsistencies in the data from a total of 1627 to 2131 (+31\%) and also discovered inconsistencies in 128 previously thought to be consistent summaries (full details can be seen in \S\ref{Benchmark}).

\paragraph{Inter Annotation Agreement.}
To validate the quality of the human annotation in our Human-LLM collaboration, we randomly sampled 150 instances and had two annotators independently review the LLM's outputs generated with the high-recall prompt. Each inconsistency predicted by the LLM was assigned with one of three categories: (1) an inconsistency present in the original data; (2) an inconsistency absent from the original data; or (3) not an inconsistency. To compute inter annotator agreement, we mapped categories (1) and (2) to True and category (3) to False, then computed inter-annotator agreement. We observed a raw agreement = 0.877 and Cohen's $\kappa$ = 0.731, indicating substantial agreement and supporting the validity of our augmentation.

\paragraph{Filtering samples.} As stated in \S\ref{Benchmark}, during the Human-LLM collaboration process the annotators encountered LLM predictions that they could not definitively confirm as correct or incorrect. To maintain dataset quality, we filtered out these ambiguous cases entirely. These instances involved cases like ambiguity, subjectivity, etc.. that made it impossible to confidently verify the model's predictions. 

For example: the LLM flagged that a summary claimed Chris Hadfield \textit{``left''} the ISS while the source text only mentioned \textit{``the time came for his departure''}. In this case the phrasing in the source text is ambiguous since it could mean that the departure time had arrived but he had not left yet, or that the time had already passed and he had departed. In the first case, the inconsistency flagged by the model is correct, but in the second it is not. This prevents us from definitively classifying the model's prediction. Another example is when the model identified an inconsistency in the summary's description of an accident as \textit{``serious''} versus the text stating the injured \textit{``required hospitalization''} the subjective definition of ``serious'' prevented a clear determination.

\subsection{Dataset Statistics}
\label{Appendix:Dataset Statistics}
\begin{figure}[H]  
    \includegraphics[width=\columnwidth]{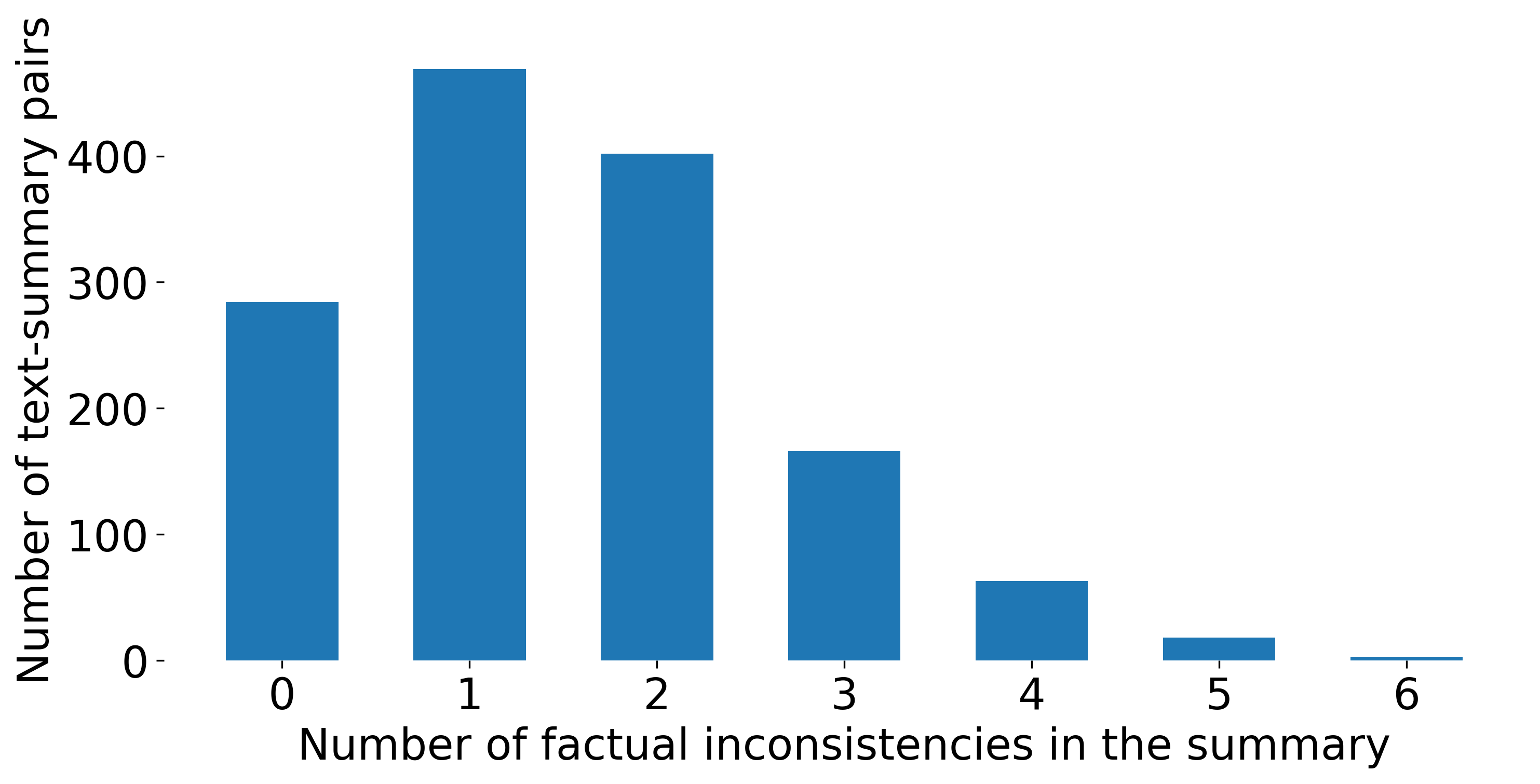}
    \caption{Number of factual inconsistencies in each summary in the final dataset.}
    \label{fig:Inconsistencies per Summary}
\end{figure}

As mentioned in \S\ref{Benchmark}, the final dataset comprises 1405 text-summary pairs with 2131 annotated factual inconsistencies. Figure~\ref{fig:Inconsistencies per Summary} presents the distribution of the number of inconsistencies per summary. Notably, roughly 45\% of summaries contain more than one inconsistency, highlighting the need for fine-grained annotation rather than a binary consistent/inconsistent label.

\section{Baselines}
\label{Appendix:Baselines}
In \S\ref{Baselines}, we introduced the three experimental setups used to evaluate LLMs for detecting factual inconsistencies. These were: (1) \textbf{E2E} - a single stage, end to end approach; (2) \textbf{Pipeline} - a multi-stage method in which we adapt FactScore \citep{factscore} to our setup and (3) \textbf{2-Step} - a two stage setup, in which the summaries are first filtered by a classifier, and then the model is prompted to detect individual errors. Below we provide additional technical details about each setup.

\subsection{E2E}
\label{Appendix:E2E}
In this setup we explored three prompting styles: \textbf{Zero-shot, Few-shot} and Chain-of-thought (\textbf{CoT}).
Since prompt phrasing can significantly impact LLM performance, we created two variants for each prompting style. For each model, we then selected the variant that achieved the highest F1 score on the development set, and used these selected prompts for test-time evaluation. All prompts variants are available in \S\ref{Appendix:Zero shot prompts} (\textbf{Zero Shot}), \S\ref{Appendix: Few shot} (\textbf{Few Shot}) and \S\ref{Appendix:Chain of Thought} (\textbf{CoT}).

\subsection{Pipeline}
\label{Appendix:Pipeline}
We adapted our Pipeline setup from FactScore \citep{factscore}. FactScore consists of two stages: (1) \textbf{decomposition} - breaking down each summary into atomic facts; and (2) \textbf{classification} - assigning each atomic fact a label of consistent or inconsistent. 
In our pipeline, the decomposition stage remained unchanged, using exactly the same prompt as in the original work. We deviate from the original paper solely in the classification stage. The original method produces a single continuous score, representing the aggregate fraction of facts classified as correct. However, our objective is not system evaluation but meta-evaluation: comparing the FactScore framework against other evaluation methodologies. This necessitated an output that allows for a direct matching between identified errors and our ground-truth annotations.
Building on the original classification prompt from FactScore, the task was changed so that instead of a single label (consistent/inconsistent), the model outputs a short description of each inconsistency it finds.
As mentioned in \S\ref{sec:representation}, an atomic fact may contain multiple inconsistencies, therefore we require the model to output a list of all the inconsistencies found in the atomic fact.
However, as illustrated in Figure \ref{fig:factscore example}, information expressed in atomic facts is not mutually exclusive, and the same content may appear in multiple facts. To remove those duplications, an additional \textbf{deduplication} stage was introduced. This stage consolidates the inconsistencies extracted from all atomic facts into a single unified list, ensuring that each unique inconsistency is represented only once. Full prompts are available at \S\ref{Appendix:Factscore}

\begin{figure}[h]
\centering
\begin{tcolorbox}[colback=gray!10, colframe=black!30, width=\linewidth, boxrule=0.5pt, arc=2mm]

\begin{minipage}{\linewidth}
\small 
\textbf{Summary}\\ The government presented new measures to stabilize prices. \\[0.5em]
\textbf{Decomposition}
\begin{enumerate}[itemsep=0pt]
    \item \textcolor{red}{The government} made a presentation.
    \item \textcolor{red}{The government} introduced new measures.
    \item The measures aim to stabilize prices.
\end{enumerate}
\textbf{Inconsistency}\\  The central bank, not the government, introduced the measures.\\[0.5em]
\textbf{Issue}\\ The inconsistency appears in (1) and (2), and causes duplicate detection.
\end{minipage}
\end{tcolorbox}
\caption{Example of decomposition to atomic facts causing duplication: when each fact in a summary is checked separately, the same inconsistency may be flagged multiple times, so de-duplication is needed.}
\label{fig:factscore example}
\end{figure}

\subsection{2-Step}
\label{Appendix:2-Step}
In the 2-Step setup, we used the CoT variant selected for each model in the E2E experiments. We evaluated three classifiers: (1) \textbf{self}, where the model itself served as a binary classifier with a chain-of-thought prompt (full prompt at \S\ref{Appendix: Binary classification prompt}) (2) \textbf{TrueTeacher}, a classifier trained for binary factual consistency detection \citep{trueteacher}; and (3) \textbf{Oracle}, where ground truth was used to filter summaries. 
The threshold for TrueTeacher was the one that maximized the F1 score on the binary summary classification task using the development set. As seen in \S\ref{Baselines}, an additional variant was implemented named \textbf{CoT\&Hint}, in which the model is explicitly instructed that the summary contains inconsistencies. For each model, this variant was created by taking its selected CoT prompt and modify it by adding a statement at the beginning indicating that the summary is inconsistent, and removing the instruction to return "None" in case no inconsistencies were found. Full \textbf{CoT\&Hint} prompts are available at \S\ref{Appendix:CoT&Hint}.

\section{Evaluation}
\label{Appendix:Evaluation}
As explained in \S\ref{Experimental setup Evaluation} and illustrated in Figure \ref{fig:process description}, we evaluate models by matching their predicted inconsistencies to ground-truth ones. This section provides additional technical details on how evaluation was conducted and explains how the evaluation pipeline was validated to ensure reliable judgments.

\paragraph{Automatic evaluation using matching.}
The judge model receives the summary, the predicted inconsistencies and the ground truth inconsistencies, and output a matching between each predicted inconsistency to a ground truth one in the form of a dictionary, as seen in Figure \ref{fig:Matching example}. A match occurs when both refer to the same inconsistent information in the summary, regardless of wording or reasoning. Each predicted inconsistency can either be matched to one ground truth inconsistency, or not matched and receive \textit{``None''}. Full matching prompt is available in \S\ref{Appendix:Matching prompt}.
Ideally, each prediction matches a different ground-truth inconsistency. However, models may repeat the same issue, either unintentionally or even to boost precision scores by repeating high-confidence predictions. This can result in multiple predictions matching the same ground-truth item.
Ideally, we would use de-duplication to remove doubles, but we found it to be less reliable for free-text outputs. Thus, instead of counting the number of matches, we count
the number of ground truth inconsistencies which were matched. This means that if a model repeats the same inconsistency multiple times, all those predictions are matched to the same ground truth and are only counted as one correct detection. This ensures that the model is rewarded correctly for detecting the inconsistency, but repeated predictions of the same inconsistency only inflate the total number of predictions without increasing the number of True Positives. In practice it means that we penalize repeated inconsistencies when we calculate the precision.

\begin{figure}[!h]
\centering
\begin{tcolorbox}[colback=gray!10, colframe=black!30, width=\linewidth, boxrule=0.5pt, arc=2mm]
\textbf{Matching input:}\\\\
\textbf{Summary:}\\ Police in Peru have clashed with squatters who have been occupying a gold mine in the north of the country.\\\\
\textbf{Predicted inconsistencies:}
\begin{itemize}
    \item[A.] The summary says the mine is in the north of the country, but the text does not mention a location.
    \item[B.] The text reports one officer killed and 10 injured, but the summary leaves this out.
    \item[C.] The summary calls it a gold mine, while the text never specifies the mineral.
\end{itemize}
\textbf{Ground truth inconsistencies:}
\begin{itemize}
    \item[A.] The summary is not correct because it adds the location being in Peru.
    \item[B.] The summary is not correct because it adds the mine being a gold mine.
    \item [C.] The summary is not correct because it adds it taking place in the north of the country.
\end{itemize}
\textbf{Matching output:}\\\\
\{"A" : "C", "B" : None , "C" : "B"\}
\end{tcolorbox}
\caption{Example of a matching output. Each predicted inconsistency is either matched to one of the ground truth inconsistencies, or not matched at all.}
\label{fig:Matching example}
\end{figure}

\paragraph{Evaluation of automatic vs ground truth matching.}
To validate our judge model, we compared its performance against that of a human judge. Human annotators were given the full set of predictions generated by the model by a specific prompt on the development set, and tasked them with performing the same matching task as the judge model, using identical instructions. The annotators completed this task across the outputs of four models (Gemini 1.5 Pro, Claude Sonnet 3.5, Llama 3.1 405B, and GPT-4o) and across all prompt types (E2E Zero-shot, E2E Few-shot, E2E CoT, and FactScore). We then compared the annotators' matches to those of the judge model and calculated recall and precision for the task. Averaging across all configurations, we obtained an average precision of 0.95 and an average recall of 0.92.

\newtcolorbox{promptbox}{
  colback=cyan!10!white,   
  colframe=black,     
  boxrule=0.5pt,    
  arc=0pt,          
  outer arc=0pt,
  left=5pt, right=5pt, top=5pt, bottom=5pt,
  fonttitle=\bfseries,
  coltitle=black,
  breakable,fontupper=\small\fontfamily{phv}\selectfont
}
\section{Error Analysis}
\label{Appendix:Error Analysis}
As mentioned in \S\ref{error analysis}, the error analysis was meant to better understand why models make mistakes. In each analysis and for each model, we sampled a random set of inconsistencies where the model was judged as incorrect and manually validated the judgement for those to exclude the possibility of judge failures.

\subsection{False Negatives Analysis}
\label{Appendix: False Negative Analysis}
In the False Negatives analysis in \S\ref{error analysis}, we examine why the model misses certain inconsistencies. Based on the categorization presented in that section and Table \ref{fig:heatmap_fn}, the most common category of undetected inconsistency is \textit{Extrinsic-Correct} - claims that are factually correct but absent from the source text. This led us to the hypothesis that such claims go undetected because they align with the model's parametric knowledge. To test this, the $\mathbf{P(True)}$ metric was introduced, which quantifies the probability the model assigns to the correctness of a specific answer to a given question. To construct question–answer pairs for computing $\mathbf{P(True)}$, we reused the 150-sample set used for categorizing the inconsistencies of GPT-4o in \S\ref{error analysis}, and selected those labeled as Extrinsic–Correct (102 samples) and Extrinsic–Wrong (17 samples). For each inconsistency, human annotators were asked to generate a question based on the source text, such that the answer would be the inconsistent information found in the summary, as can be seen in Figure~\ref{fig:armstrong-example}. Questions were derived exclusively from the source text, so the model has the same information it had when it originally failed to flag the inconsistency.

\begin{figure}[h!]
\centering
\begin{tcolorbox}[colback=gray!10, colframe=black!30, width=\linewidth, boxrule=0.5pt, arc=2mm]

\begin{minipage}{\linewidth}
\small 
\textbf{Text}\\ Neil Armstrong was an American astronaut and the first person to walk on the Moon during NASA's Apollo 11 mission. He famously said, "That's one small step for man, one giant leap for mankind." Armstrong was also a test pilot and aerospace engineer. \\[0.5em]
\textbf{Summary}\\ Neil Armstrong was the first person to walk on the Moon on \textcolor{red}{20, July 1969.}\\[0.5em]
\textbf{Question}\\ On what date did Neil Armstrong walk the moon for the first time?\\[0.5em]
\textbf{Answer}\\ July 20th 1969.
\end{minipage}
\end{tcolorbox}

\caption{Example of question generation for P(True)}
\label{fig:armstrong-example}
\end{figure}

To evaluate \(P(\text{true})\) on GPT-4o, we followed the following procedure:

\begin{enumerate}
    \item \textbf{Bias Elimination} We ran the prompt in two configurations:
    \begin{itemize}
        \item Exactly as shown below.
        \item With the answer options swapped (option A marked incorrect and option B correct).
    \end{itemize}
    \item \textbf{Repeated sampling} Because the model's responses vary slightly between runs, we executed each prompt configuration 10 times and averaged the resulting probabilities.
\end{enumerate}
All sampling was done with temperature of 1.
The prompt used to estimate \(P(\text{true})\) was adapted from \citep{insideout}, and is shown below.
\begin{promptbox}
\textbf{System Prompt:}\\\\
Your job is to evaluate if a proposed answer to a question is correct.\\\\
\textbf{Prompt:}\\\\
Here is the question and the proposed answer.\\\\
Question: <question>\\
Answer: <answer>\\\\
Is the proposed answer:\\
A: CORRECT\\
B: INCORRECT\\
Just return the letters "A" or "B", with no text around it.
\end{promptbox}
\subsection{False Positives Analysis}
\begin{table*}[!t]
\centering
\colorbox{gray!10}{\begin{minipage}{1\textwidth}
\vspace{0.5em}
\textbf{Text:}\\
\small A Japanese spacecraft successfully landed on an asteroid and collected samples in a groundbreaking mission. The Hayabusa2 probe touched down, fired a projectile to gather material, and later returned the samples to Earth. Scientists aimed to study the asteroid's composition to learn more about the origins of the solar system. The mission was hailed as a major achievement in space exploration.
\vspace{0.5em}\\
\textbf{Summary:}\\
\small Japan's Hayabusa2 spacecraft landed on an asteroid, collected samples, and has returned them to Earth for solar system research.
\vspace{0.5em}
\end{minipage}}
\vspace{0.5em}\\
\resizebox{\textwidth}{!}{
\begin{tabular}{>{\centering\arraybackslash}p{2cm} >{\arraybackslash}p{4cm} p{4cm} p{3.5cm} p{4.5cm}}
\hline
\textbf{Category} & \textbf{\centering Definition} & \textbf{\centering Model Output} & \textbf{\centering  Evidence} & \textbf{\centering Explanation} \\
\hline
Overlooked Info & \small The model fails to recognize information that is explicitly stated in the source text, resulting in an incorrect detection of an inconsistency. &
\small The text never claimed the samples have returned to Earth. &
\small ...The
Hayabusa2 probe touched down, fired a projectile to gather material, and \textcolor{red}{later returned the samples to Earth}...&
\small The model simply misses this part of the text. \\
\hline
Missed Deduction & \small The model fails to recognize a fact that can be directly deduced from the text, and classifies the information as wrong. &
\small The text does not state that Hayabusa2 is from Japan. It states an unnamed Japanese spacecraft landed on an asteroid and later discusses the Hayabusa2 landing. &
\small A \textcolor{red}{Japanese spacecraft} successfully landed...\textcolor{red}{The
Hayabusa2 probe} touched down &
\small The model failed to recognize that both references describe the same spacecraft. \\
\hline
Omission & \small The model considers information missing from the summary as an inconsistency, even if the summary is still factually consistent. &
\small The summary fails to mention it was a major achievement for space exploration, which is a key detail in the text. &
\small ...\textcolor{red}{The mission was hailed as a major achievement in space exploration.} &
\small The model views a lack of a specific detail (it being a major achievement) as an inconsistency, even if the summary is factually correct. \\
\hline
Overly Literal & \small The model identifies superficial changes in wording as inconsistencies. &
\small The summary claims the Hayabusa2 spacecraft landed, but the text explicitly calls it "The Hayabusa2 probe". &\small A Japanese \textcolor{red}{spacecraft} ... The Hayabusa2 \textcolor{red}{probe} touched down... &
\small The model views the use of slightly different (but still supported) terminology of "spacecraft" vs "probe" as an inconsistency. \\
\hline
Invented & \small The model claims there is inconsistency, but the information it mentions as inconsistent is not in the text or summary. &
\small The summary implies the landing is recent ("has returned"), but the text does not specify when it happened. &
\small ..and later \textcolor{red}{returned} the samples to Earth &
\small The model treats 'has returned' as recent, though present perfect can describe distant events.\\
\hline
\end{tabular}
}
\captionsetup{
  skip=0pt,
  belowskip=0pt
}
\caption{Categorization of False Positives. Above the table is a text, and a factually consistent summary of it. The table presents the categories of false positives, each demonstrated with an example based on this text-summary pair. Each row shows a case where the model flagged a factual inconsistency, even though it does not exist. For each example, the table also provides an explanation of why it was assigned to that particular category.}
\label{tab:false-postives-categories}
\end{table*}

In the False Positives analysis discussed in \S\ref{error analysis}, we examine cases where the models predicted that certain information was an inconsistency, even though this was not true. While the categories of false positives are specified in that section, here we provide a more detailed illustration of each category, as shown in Table~\ref{tab:false-postives-categories}.
\section{Prompts}

\subsection{High Recall Prompt}
\label{Appendix:High Recall Prompt}
This is the High recall prompt mentioned in \S\ref{Benchmark} and \S\ref{Appendix:Human-LLM Collaboration}, used to maximize the amount of inconsistencies the model detects, to later be filtered by humans to uncover additional inconsistencies.
\begin{promptbox}
I will give you a text and a summary. The summary is factually inconsistent with respect to the text. Your task is to identify and explain all factual inconsistencies in the summary. A factual inconsistency is any information in the summary that cannot be verified by the original text. \\\\
In your explanations focus only on factual inconsistencies, and no other types of mistakes.\\\\
List all the facts you found to be inconsistent with the text. Each inconsistent fact should appear separately. List each inconsistent fact as briefly as possible, along with a short description. The fact should be the minimal span of words which is not supported by the text, and represent the mistake. The description should be brief and concise, and describe exactly what is the mistake, with no ambiguity.\\\\
The following are examples of texts, summaries and the corresponding lists of inconsistent facts.\\\\
\textbf{Text:}\\
A 27-year-old man and a woman, 32, were detained after the 60-year-old victim's body was found at the Forest Gate house, early on Christmas Day. Four people escaped from the house on Field Road before firefighters arrived just before 04:45 GMT. A post-mortem test showed the victim had died from burns and the inhalation of fumes, the Met Police said. Fire crews found his body on the ground-floor of the two-storey house. Police believe "the fire was started deliberately" and say they believe they know who the victim was, but formal identification has not yet been made. Twenty one firefighters and four engines tackled the blaze, which was brought under control after about two-and-a-half hours. Det Ch Insp Steve McCabe said: "I need to hear from anyone who saw anything suspicious in Field Road and the surrounding area in the early hours of Christmas Day.\\\\
\textbf{Summary:}\\
Two people have been arrested on suspicion of murder after a man died in a house fire in east London.\\\\
A.\\
Fact:
arrested on suspicion of murder\\
Description:
The summary states the people have been arrested on suspicion of murder, but the source text does not state the charges against them.\\
B.\\
Fact:
east London\\
Description:
The summary states that the fire took place in east London, but this information does not appear in the text.\\\\ 
\textbf{Text:}\\
The Woodland Trust wants to buy the land at Llennyrch woodland. Natural Resources Wales (NRW) has given £50,000 but a further £750,000 is needed and a campaign will be launched on Tuesday at the National Eisteddfod in Meifod, Powys. The charity said the area has been called a "Celtic rainforest" and it wants to improve wildlife on the site. NRW chief executive Dr Emyr Roberts said: "This is a fantastic opportunity to bring the whole woodland area under conservation management." The total cost of the project is £1.5m and the rest of the cost will be met by money left to the Woodland Trust.\\\\
\textbf{Summary:}\\
A campaign has been launched to raise £1m to buy 1,000 acres of woodland in Carmarthenshire.\\\\
A.\\
Fact:
1000 acres\\
Description:
The summary states they want to buy 1000 acres of woodland, but acreage is not mentioned in the source text.\\
B.\\
Fact:
£1m\\
Description:
The summary states the campaign wants to raise £1m, but the source text says the campaign want to raise an additional £750,000.\\
C.\\
Fact:
Carmarthenshire\\
Description:
The summary states the woodland is in Carmarthenshire, but the source text says it's in Llennyrch.\\
D.\\
Fact:
has been\\
Description:
The summary states the campaign has been launched, but the source text says it will be launched.\\\\
Here is a new example: \\\\
Text: <Text>\\
Summary: <Summary>\\
Output the list of inconsistent facts and explanations in the same format as in the examples above.
\end{promptbox}

\subsection{Detection Prompts}
\label{Appendix:Detection Prompts}
Those are some of the prompts used to perform the detection of factual inconsistencies in \S\ref{sec:exp_setup}. We have 2 variants for each \textbf{E2E} prompt, but we decided to show only one of each for brevity.
\vspace{0.5em}

\subsubsection{Zero Shot}
\label{Appendix:Zero shot prompts}
\textbf{Prompt 1:}
\begin{promptbox}
I will give you a text and a summary. Your task is to identify and explain all factual inconsistencies in the summary. A factual inconsistency is any information in the summary that cannot be verified by the original text.\\\\
Focus only on factual inconsistencies, and no other types of mistakes, such as omission.\\\\
List all the facts you found to be inconsistent with the text. Mark each inconsistent fact with letters A, B, C, etc., in sequential order. Each inconsistent fact should appear separately. List each inconsistent fact as a short description of the inconsistency. The description should be brief and concise.\\\\
If there are no factual inconsistencies in the summary, output None.\\\\
Text: <Text>\\
Summary:<Summary>
\end{promptbox}
\textbf{Prompt 2:}
\begin{promptbox}
I will give you a text and a summary. Your task is to identify all factual inconsistencies and briefly describe each one. If there are no factual inconsistencies, return “None.”\\\\
A “factual inconsistency” is any detail in the summary that contradicts or cannot be verified by the text.
Focus only on these inconsistencies; do not address omissions or any other errors.\\\\
List each inconsistency separately, labeling them with letters (A, B, C, etc.).
For each labeled item, provide a short description of what the inconsistency is.
Keep your descriptions concise and only address factual inconsistencies.\\\\
Text: <Text>\\
Summary: <Summary>
\end{promptbox}
\subsubsection{Few Shot}
\label{Appendix: Few shot}
\textbf{Prompt 1:}
\begin{promptbox}
I will give you a text and a summary. Your task is to identify and explain all factual inconsistencies in the summary. A factual inconsistency is any information in the summary that cannot be verified by the original text. \\\\
Focus only on factual inconsistencies, and no other types of mistakes, such as omission.\\\\
List all the facts you found to be inconsistent with the text. Each inconsistent fact should appear separately. List each inconsistent fact as a short description of the inconsistency. The description should be brief and concise.\\\\
If there are no factual inconsistencies in the summary, return None.\\\\
The following are examples of texts, summaries and the corresponding lists of such facts.\\\\
\textbf{Text:}\\
A 27-year-old man and a woman, 32, were detained after the 60-year-old victim's body was found at the Forest Gate house, early on Christmas Day. Four people escaped from the house on Field Road before firefighters arrived just before 04:45 GMT. A post-mortem test showed the victim had died from burns and the inhalation of fumes, the Met Police said. Fire crews found his body on the ground-floor of the two-storey house. Police believe "the fire was started deliberately" and say they believe they know who the victim was, but formal identification has not yet been made. Twenty one firefighters and four engines tackled the blaze, which was brought under control after about two-and-a-half hours. Det Ch Insp Steve McCabe said: "I need to hear from anyone who saw anything suspicious in Field Road and the surrounding area in the early hours of Christmas Day.\\\\
\textbf{Summary:}\\
Two people have been arrested on suspicion of murder after a man died in a house fire in east London.\\\\
A.\\
Description:
The summary states the people have been arrested on suspicion of murder, but the source text does not state the charges against them.\\
B.\\
Description:
The summary states that the fire took place in east London, but this information does not appear in the text.\\\\
\textbf{Text:}\\
The Woodland Trust wants to buy the land at Llennyrch woodland. Natural Resources Wales (NRW) has given £50,000 but a further £750,000 is needed and a campaign will be launched on Tuesday at the National Eisteddfod in Meifod, Powys. The charity said the area has been called a "Celtic rainforest" and it wants to improve wildlife on the site. NRW chief executive Dr Emyr Roberts said: "This is a fantastic opportunity to bring the whole woodland area under conservation management." The total cost of the project is £1.5m and the rest of the cost will be met by money left to the Woodland Trust.\\\\
\textbf{Summary:}\\
A campaign has been launched to raise £1m to buy 1,000 acres of woodland in Carmarthenshire.\\\\
A.\\
Description:
The summary states they want to buy 1000 acres of woodland, but acreage is not mentioned in the source text.\\
B.\\
Description:
The summary states the campaign wants to raise £1m, but the source text says the campaign wants to raise an additional £750,000.\\
C.\\
Description:
The summary states the woodland is in Carmarthenshire, but the source text says it's in Llennyrch.\\
D.\\
Description:
The summary states the campaign has been launched, but the source text says it will be launched.\\\\
\textbf{Text:}\\
The 14-month old tabby and white called Pumbaa was found bleeding in a Peterborough alleyway on Saturday. The stab wound was so deep the vet was unable to operate before Pumbaa died. A second cat - Mischief - was shot by an air rifle in an area near to where Pumbaa was stabbed, according to the RSPCA. It is unclear whether the two incidents are linked. RSPCA inspector Justin Stubbs said: "These were two shocking and completely senseless attacks." Pumbaa's owner, Kirsty Cracknell, 29, of Croyland Road, said: "I am utterly devastated about Pumbaa - he was such a soppy little mummy's boy. I just keep expecting him to jump through the window. "What particularly breaks my heart is that I think he must have been on his way home to me, considering where he was found."\\\\
\textbf{Summary:}\\
A cat has been stabbed to death in what the RSPCA described as a "senseless attack".\\\\
None\\\\
Here is a new example. \\\\
Text: <Text>\\
Summary:<Summary>\\\\
Output the list in the same format as in the examples above.
\end{promptbox}
\textbf{Prompt 2:}
\begin{promptbox}
Task: Identify Inconsistencies in Summary Texts\\\\
Objective: The given summary that may contain factual inconsistencies. Your task is to critically analyze and document these inconsistencies by comparing the summary to the original text.
If there are no factual inconsistencies in the summary, return None.\\\\
Definition of Factual Inconsistency:
\begin{itemize}
    \item[-] Any statement in the summary that cannot be directly verified or supported by the original text.
    \item[-] Discrepancies in names, locations, numbers, events, or specific claims. 
\end{itemize}
Evaluation Criteria:
\begin{enumerate}
    \item Identify each distinct factual inconsistency.
    \item Describe each inconsistency briefly and precisely.
    \item Focus solely on factual discrepancies, not stylistic or structural differences.
    \item Do not comment on omissions or missing information.
\end{enumerate}
The following are examples of texts, summaries and the corresponding lists of such facts.\\\\
\textbf{Text:}\\
A 27-year-old man and a woman, 32, were detained after the 60-year-old victim's body was found at the Forest Gate house, early on Christmas Day. Four people escaped from the house on Field Road before firefighters arrived just before 04:45 GMT. A post-mortem test showed the victim had died from burns and the inhalation of fumes, the Met Police said. Fire crews found his body on the ground-floor of the two-storey house. Police believe "the fire was started deliberately" and say they believe they know who the victim was, but formal identification has not yet been made. Twenty one firefighters and four engines tackled the blaze, which was brought under control after about two-and-a-half hours. Det Ch Insp Steve McCabe said: "I need to hear from anyone who saw anything suspicious in Field Road and the surrounding area in the early hours of Christmas Day.\\\\
\textbf{Summary:}\\
Two people have been arrested on suspicion of murder after a man died in a house fire in east London.\\\\
\textbf{Inconsistencies:}\\
A.\\
Description:
The summary states the people have been arrested on suspicion of murder, but the source text does not state the charges against them.\\
B.\\
Description:
The summary states that the fire took place in east London, but this information does not appear in the text.\\\\
\textbf{Text:}\\
The Woodland Trust wants to buy the land at Llennyrch woodland. Natural Resources Wales (NRW) has given £50,000 but a further £750,000 is needed and a campaign will be launched on Tuesday at the National Eisteddfod in Meifod, Powys. The charity said the area has been called a "Celtic rainforest" and it wants to improve wildlife on the site. NRW chief executive Dr Emyr Roberts said: "This is a fantastic opportunity to bring the whole woodland area under conservation management." The total cost of the project is £1.5m and the rest of the cost will be met by money left to the Woodland Trust.\\\\
\textbf{Summary:}\\
A campaign has been launched to raise £1m to buy 1,000 acres of woodland in Carmarthenshire.\\\\
\textbf{Inconsistencies:}\\
A.\\
Description:
The summary states they want to buy 1000 acres of woodland, but acreage is not mentioned in the source text.\\
B.\\
Description:
The summary states the campaign wants to raise £1m, but the source text says the campaign wants to raise an additional £750,000.\\
C.\\
Description:
The summary states the woodland is in Carmarthenshire, but the source text says it's in Llennyrch.\\
D.\\
Description:
The summary states the campaign has been launched, but the source text says it will be launched.\\\\
\textbf{Text:}\\
The 14-month old tabby and white called Pumbaa was found bleeding in a Peterborough alleyway on Saturday. The stab wound was so deep the vet was unable to operate before Pumbaa died. A second cat - Mischief - was shot by an air rifle in an area near to where Pumbaa was stabbed, according to the RSPCA. It is unclear whether the two incidents are linked. RSPCA inspector Justin Stubbs said: "These were two shocking and completely senseless attacks." Pumbaa's owner, Kirsty Cracknell, 29, of Croyland Road, said: "I am utterly devastated about Pumbaa - he was such a soppy little mummy's boy. I just keep expecting him to jump through the window. "What particularly breaks my heart is that I think he must have been on his way home to me, considering where he was found."\\\\
\textbf{Summary:}\\
A cat has been stabbed to death in what the RSPCA described as a "senseless attack".\\\\
\textbf{Inconsistencies:}\\
None\\\\
Here is a new example:\\
Text: <Text>\\
Summary: <Summary>\\
Output the list in the same format as in the examples above.
\end{promptbox}
\subsubsection{Chain of Thought}
\label{Appendix:Chain of Thought}
\textbf{Prompt 1:}
\begin{promptbox}
I will give you a text and a summary. Your task is to identify all the facts in the summary that cannot be verified using the text and clearly describe each factual inconsistency. Think step by step.\\\\
Text: <Text>\\
Summary:<Summary>\\\\
At the end of your response, under Final Output:, output each unverifiable fact with a clear description of the factual inconsistency. Mark each unverifiable fact with letters A, B, C, etc., in sequential order.
\end{promptbox}
\textbf{Prompt 2:}
\begin{promptbox}
I will give you a text and a summary.
Your task is to identify all factual inconsistencies by following these steps:\\
Examine the summary carefully and break it down into distinct factual statements.\\
For each statement, check whether it is:\\
Contradicted by the text (directly conflicts with the text).\\
Not verifiable from the text (lacks necessary support in the text).\\
For each inconsistency, describe exactly what is incorrect in the summary.\\\\
Text: <Text>\\
Summary: <Summary>\\\\
Think step by step, and at the end of your response, under "Final Output:", provide a clear description of each mistake.\\
Mark each inconsistency with letters A, B, C, etc.\\
Focus only on what is wrong in the summary.\\
If no factual inconsistencies were identified, under "Final Output:" return None.
\end{promptbox}
\subsubsection{CoT\&Hint}
These prompts are based on the CoT prompts in \S\ref{Appendix:Chain of Thought}, but include an explicit statement that the summary contains inconsistencies. This addition is highlighted in the prompts below. Furthermore, in case there was an instruction in the prompt to return "None" in case of no inconsistencies, it was removed.

\label{Appendix:CoT&Hint}
\textbf{Prompt 1:}
\begin{promptbox}
I will give you a text and a summary. \textbf{The summary is factually inconsistent with respect to the text.} Your task is to identify all the facts in the summary that cannot be verified using the text and clearly describe each factual inconsistency. Think step by step.\\\\
Text: <Text>\\
Summary:<Summary>\\\\
At the end of your response, under Final Output:, output each unverifiable fact with a clear description of the factual inconsistency. Mark each unverifiable fact with letters A, B, C, etc., in sequential order.
\end{promptbox}
\textbf{Prompt 2:}
\begin{promptbox}
I will give you a text and a summary. \textbf{The summary contains factual inconsistencies with respect to the text.}
Your task is to identify all factual inconsistencies by following these steps:\\
Examine the summary carefully and break it down into distinct factual statements.\\
For each statement, check whether it is:\\
Contradicted by the text (directly conflicts with the text).\\
Not verifiable from the text (lacks necessary support in the text).\\
For each inconsistency, describe exactly what is incorrect in the summary.\\\\
Text: <Text>\\
Summary: <Summary>\\\\
Think step by step, and at the end of your response, under "Final Output:", provide a clear description of each mistake.\\
Mark each inconsistency with letters A, B, C, etc.\\
Focus only on what is wrong in the summary.
\end{promptbox}
\subsubsection{FactScore}
\label{Appendix:Factscore}
\textbf{Decomposition prompt:}
\vspace{0.5em}

\noindent
We used the original prompt given in \citep{factscore}.

\vspace{0.5em}

\noindent
\textbf{Detection prompt:}
\begin{promptbox}
Is the following atomic fact factually consistent with the text?
A factual inconsistency is any information in the summary that cannot be verified by the original text. Respond with "yes" or "no".\\
If and only if the answer is "no", list all the inconsistencies you identified. \\
Each inconsistency must appear separately and begin with a bullet mark (-). Do not group multiple issues in one bullet. \\
Each description must clearly identify what the atomic fact claims and explain what specific detail is missing, wrong, or not supported in the text.\\
Avoid vague statements like "this information can not be verified from the text."\\
Each description should be brief and concise.
\end{promptbox}
\textbf{Deduplication prompt:}
\begin{promptbox}
You will be given a summary and a list of factual inconsistency descriptions referring to that summary. Your task is to create a new list that removes duplicate descriptions of the same factual inconsistency. Follow these instructions carefully:\\
Identify which descriptions refer to the same incorrect piece of information in the summary, even if the wording or reasoning differs.\\
From each group of descriptions referring to the same inconsistency, keep only the one that most clearly specifies what the exact inconsistency is. Discard the rest.\\
Do not drop descriptions for any reason beside merging.\\
Do not rewrite or modify any description. Simply select the best one among duplicates and preserve it as-is.\\
If no duplicates are found, do not make any changes to the input list.\\
Enumerate the final list using letters: A., B., C., etc.
\end{promptbox}
\subsection{Binary classification prompt}
\label{Appendix: Binary classification prompt}
This is the prompt referenced in \S\ref{sec:exp_setup} for the 2-Step setup, labeled as \textbf{Self}, where the model first acts as a binary classifier before identifying individual inconsistencies.
\begin{promptbox}
Decide if the following summary is factually consistent with the corresponding text. Note that consistency means all information in the summary is supported by the text.\\\\
Text: <Text>\\
Summary: <Summary>\\\\
Explain your reasoning step by step and end your response with final answer: yes or no only
\end{promptbox}
\subsection{Matching prompt}
This is the judgment prompt, referenced in \S\ref{sec:exp_setup} and \S\ref{Appendix:Evaluation}, used to evaluate a model's predictions by comparing them with the ground truth.
\label{Appendix:Matching prompt}
\begin{promptbox}
You are tasked with analyzing two lists of descriptions: one labeled as the Predicted Output and the other as the Gold Label. The goal is to determine whether each description in the Predicted Output is specific and matches a description in the Gold Label.\\\\
Input Format\\
Summary: A text containing the factual inconsistencies to be evaluated.\\
Predicted Output: A set of descriptions labeled with identifiers (e.g., A, B, C).\\
Gold Label: A set of descriptions labeled with identifiers (e.g., A, B, C).\\\\
Detailed Instructions:\\\\
For each description in the Predicted Output:\\\\
Compare the exact description of the inconsistency to items in the Gold Label.\\
A match occurs if:\\
Both descriptions identify the same fact from the summary as being wrong, regardless of why that fact is considered incorrect.\\
Ensure the match is based on the specific information addressed in both descriptions.\\
The description in the Predicted Output is not ambiguous - it clearly refers to one specific fact from the summary. Vague or overly broad descriptions should not be matched.\\
Each description in the Predicted Output can match at most one description in the Gold Label.\\\\
If the Predicted Output is empty (no descriptions were given, or any other input is given), return an empty JSON object ({}).\\
If the Gold Label is empty, return a JSON object with the keys of the Predicted Output and null value for each.\\\\
Summary: <Summary>\\
Gold Label: <Gold Label>\\
Predicted Output: <Predicted Output>\\\\
Think about the matching step by step and output a JSON object:\\\\
For each description in the Predicted Output, provide:\\
A key corresponding to the predicted description's identifier (e.g., A, B, C, etc.).\\
If a match is found, set the value to the identifier from the Gold Label that matches it.\\
If no match is found, set the value to null.
\end{promptbox}

\end{document}